\documentclass{article}
\usepackage{arxiv}
\usepackage[utf8]{inputenc} 
\usepackage[T1]{fontenc}    
\usepackage{hyperref}       
\usepackage{url}            
\usepackage{booktabs}       
\usepackage{amsfonts}       
\usepackage{nicefrac}       
\usepackage{microtype}      
\usepackage{lipsum}
\usepackage{graphicx}
\graphicspath{ {./images/} }
\usepackage{subcaption}
\usepackage{multicol}

\usepackage{multirow} %
\newtheorem{remark}{Remark}

\title{On the robustness of ChatGPT in teaching Korean Mathematics}

\author{ 
    Phuong-Nam Nguyen \\
    Faculty of Information Technology\\
    College of Technology,\\
    National Economics University\\
    Hanoi, 100000, Vietnam \\
    \texttt{namnp@neu.edu.vn} \\
    \And
    Quang Nguyen-The \\
    Faculty of Information Technology\\
    College of Technology,\\
    National Economics University\\
    Hanoi, 100000, Vietnam \\
    \texttt{nguyenthequang.neu@gmail.com} \\
    \And 
    An Vu-Minh \\    Faculty of Mathematical Economics\\
    College of Technology\\
    National Economic University\\
    Hanoi, 100000, Vietnam \\
    \texttt{vuxminhan@gmail.com} \\
    \And
    Diep Anh Nguyen \\
    Faculty of Information Technology\\
    College of Technology,\\
    National Economics University\\
    Hanoi, 100000, Vietnam \\
    \texttt{diepanhnguyen2807@gmail.com} \\
    \And 
    Xuan-Lam Pham \\
    Faculty of Information Technology\\
    College of Technology,\\
    National Economics University\\
    Hanoi, 100000, Vietnam \\
    \texttt{lamxp@neu.edu.vn} \\
}

\begin{document}
\maketitle

\begin{abstract}

\noindent \textbf{Background:} ChatGPT is an Artificial Intelligence that enables the revolution of many fields, which promises the path toward personalized education. However, the AI models fall short when tackling questions not in English. Thus, we believe investigating the reliability and robustness of such models in teaching and solving multilingual science questions is the first step toward fully adopting AI for personalized education. 

\noindent \textbf{Approach:} We use Korean mathematics questions as a validated dataset, which includes $586$ questions. Other than testing the accuracy of ChatGPT, we evaluate the model's effectiveness in rating mathematics questions using eleven criteria. We also perform topic analysis, suggesting effective use of the models.

\noindent \textbf{Results:} Out of $586$ questions, ChatGPT achieves about $66.72\%$ of accuracy (correctly answer $n=391/586$ questions). Besides, ChatGPT's rating ability is substantially good and consistent with education theory and test taker's perspectives.

\noindent \textbf{Conclusion:} The results highlight both the potential and limitations of ChatGPT in multilingual educational settings. While the model demonstrates a reasonable degree of accuracy (66.72\%) in solving Korean mathematics questions, this also indicates room for improvement in handling non-English contexts. Its strong performance in question rating, which aligns with established educational theories and user perspectives, underscores its utility for assessment and content analysis. These findings suggest that ChatGPT can be valuable in personalized education, particularly when supported by domain-specific optimizations and multilingual training enhancements. Future work should address linguistic biases, improve accuracy across diverse languages, and integrate insights from topic analysis to enhance the model's reliability and effectiveness in diverse educational environments. This will pave the way for the broader adoption of AI in global and inclusive education systems.
\end{abstract}




\section{Introduction}
ChatGPT, an AI technology powered by Large Language Models (LLMs), has the potential to revolutionize various fields, including Natural Language Processing (NLP), computer vision, molecular analysis, and educational technology. In its commercialization, OpenAI reports that the model versions—GPT-4, GPT-4 (no vision), and GPT-3.5—achieve impressive performance in standardized tests. For instance, they rank in the 93rd, 93rd, and 87th percentiles, respectively, on the SAT evidence-based reading and writing section (undergraduate entrance). However, performance declines in the SAT mathematics section, with GPT-4 scoring in the 89th percentile and GPT-3.5 dropping to the 70th percentile. A similar trend is observed in higher-level assessments. On the GRE verbal reasoning section (graduate entrance), the models achieve scores ranging from the 63rd to 99th percentile. Conversely, their performance in the GRE quantitative reasoning section is less consistent, spanning from the 25th to 80th percentiles. These results highlight a notable research gap: the effective resolution of mathematical problems remains a significant challenge for current AI models. Another limitation of these AI technologies becomes evident in non-English examinations. A study by \cite{siebielec2024assessment} assesses ChatGPT-3.5's performance on Poland's Medical Final Examination, which tests comprehensive medical knowledge. Analyzing 980 questions from 2022–2024, the model achieved an average accuracy of ~60\%, falling short of human examinees. Similarly, \cite{hsieh2024evaluating} evaluates ChatGPT-3.5 and GPT-4 on 1,375 questions from Taiwan's Plastic Surgery Board Examination. GPT-4 outperformed GPT-3.5, achieving a 59\% correct response rate compared to GPT-3.5’s 41\%. Notably, GPT-4 passed five of the eight exams, while GPT-3.5 failed all. These findings underscore the current limitations of AI in specific domains, particularly in mathematics and non-English assessments, while highlighting its strengths in verbal and reasoning tasks.

LLM like ChatGPT have transformed the way individuals interact with artificial intelligence (AI) in educational contexts. Mathematics, being a discipline characterized by precision, logic, and structure, presents unique challenges for AI systems. The efficacy of ChatGPT in solving mathematics questions has been explored in contexts ranging from basic arithmetic to advanced calculus and problem-solving. \cite{mukherjee2008review} reviews the progress in using machines to understand and solve mathematical problems described in natural language, a research area dating back to the 1960s. It provides a comprehensive technical overview of systems and approaches developed for various domains such as algebra, geometry, physics, and mechanics. \cite{wardat2023chatgpt} explores the perspectives of students and educators on using ChatGPT for teaching mathematics, revealing both its strengths, such as improved math capabilities and positive public discourse, and its limitations, including a lack of deep understanding in geometry and challenges in correcting misconceptions. It highlights the need for further research to ensure the safe and effective integration of AI tools like ChatGPT into mathematics education. \cite{frieder2024mathematical} evaluates the mathematical capabilities of ChatGPT (January 2023 iterations) and GPT-4 using novel datasets, GHOSTS, and miniGHOSTS, which cover graduate-level mathematics and assess multiple dimensions of reasoning. While GPT-4 performs well as a mathematical assistant for querying facts and undergraduate-level tasks, it struggles with graduate-level difficulty, with overall performance falling below that of a graduate student, challenging the media's optimistic portrayal of these models' exam-solving abilities. \cite{daher2024use} examines the language used by ChatGPT in solving quadratic equation problems, employing a functional grammar framework to analyze its responses. While ChatGPT generally supports learning by explaining problem-solving steps, its occasional mathematical errors and reliance on non-material processes suggest that its use should be accompanied by teacher guidance to prevent reinforcing misconceptions in students. The technical report\cite{kashefi2023chatgpt} evaluates ChatGPT's ability to program numerical algorithms, including generating code, debugging, completing missing parts, and parallelizing code across various programming languages. While ChatGPT shows promise in programming tasks for mathematical problems and scientific machine learning applications, it faces challenges such as handling singular matrices, array size mismatches, and long code interruptions, highlighting the need for further improvements in the model. In a more complex mathematical concept, \cite{liu2024chatgpt} explores the use of ChatGPT to bridge the gap between theoretical topological concepts and their practical implementation in computational topology. By guiding ChatGPT to generate and validate functional codes for tasks such as computing Betti numbers, Laplacian matrices, and homology persistence, the study demonstrates how mathematicians without coding experience can effectively leverage AI for computational tasks, advancing the practical application of topological data analysis. 

\cite{peng2024multimath} introduces MultiMath-7B, a multimodal large language model that integrates mathematical reasoning with visual inputs like diagrams, charts, and function plots. Trained through a four-stage process, it outperforms existing open-source models on multimodal and text-only mathematical benchmarks and is supported by the novel MultiMath-300K dataset, which includes diverse mathematical tasks from K-12 levels with image captions and step-by-step solutions. \cite{deng2024enhancing} introduces Self-Training on Image Comprehension (STIC), a self-training approach designed to improve the image comprehension capabilities of large vision-language models (LVLMs) without requiring labeled data. By allowing the model to generate preferred and dis-preferred image descriptions and leveraging a small portion of existing instruction-tuning data, STIC achieves a 4.0\% performance improvement on seven benchmarks while using 70\% less supervised fine-tuning data, demonstrating the potential to utilize large amounts of unlabeled images for model enhancement. \cite{zhang2025mathverse} introduces MathVerse, a comprehensive visual math benchmark designed to evaluate the true capabilities of multi-modal large language models (MLLMs) in interpreting visual diagrams for mathematical reasoning. By collecting 2,612 high-quality math problems with diagrams and transforming them into six distinct versions, MathVerse enables a nuanced assessment of MLLMs' understanding of visual content, complemented by a Chain-of-Thought (CoT) evaluation strategy using GPT-4(V) to analyze and score reasoning steps.

In this article, we aim to answer two research questions (RQs):
\begin{description}
    \item[RQ1] What is the accuracy of ChatGPT in solving mathematics in a multilingual context, specifically in Korean?
    \item[RQ2] What is the effectiveness of ChatGPT in rating and categorizing mathematical questions at the university entrance level?
\end{description}
These RQs are important because they address the practical application of ChatGPT in real-world educational contexts. RQ1 explores ChatGPT's ability to solve mathematical problems in Korean, which is crucial for understanding its effectiveness in non-English languages, while RQ2 evaluates its ability to rate and categorize university-level math questions, assessing whether ChatGPT can be used as a reliable tool for academic testing and evaluation. These questions help determine ChatGPT's potential role in multilingual education and its utility in academic assessments.

\section{Study design}
In this research, we use The CSAT (College Scholastic Ability Test), which is a standardized exam in South Korea that plays a critical role in university admissions, particularly for undergraduate programs. It is a comprehensive test that evaluates students' knowledge across various subjects, with a mathematics section designed to assess their proficiency in advanced mathematical concepts and problem-solving skills. The mathematics portion of the CSAT covers a wide range of topics, including algebra, geometry, calculus, and statistics, and it is known for its challenging questions that require a deep understanding of mathematical principles. We use a validated dataset consisting of $586$ Korean mathematics questions from the CSAT, which allows for a detailed evaluation of ChatGPT’s performance in solving mathematical problems in a multilingual context. In addition to testing the accuracy of ChatGPT's responses, we assess the model's effectiveness in rating and categorizing the questions according to eleven criteria, including difficulty level, topic coverage, and clarity. Furthermore, we perform a topic analysis to explore how the model can be effectively used to support mathematics education, providing insights into its potential applications in academic settings.

To handle RQ1, we follow these step-by-step instructions:
\begin{description}
    \item[Step 1: Data preparation] We collect CSAT mathematics questions from official test in \texttt{pdf} format. Then, we preprocess the dataset to ensure consistency in formats (\texttt{.tex} and \texttt{.md}).
    \item[Step 2: Model selection] We set up ChatGPT by loading its pre-trained model and configuring it for zero-shot evaluation.
    \item[Step 3: Model testing] We input each mathematics question from the test set into ChatGPT. Then we collect the model's responses and compare them to the ground truth answers provided in the dataset. Finally, we evaluate the accuracy of the model’s responses based on the official answer sheets provided by the test distributors.
\end{description}

For RQ2, we follow the below steps:
\begin{description}
    \item[Step 1: Data preparation] We use the same dataset of $586$ CSAT mathematics questions. We design a framework to evaluate the model’s performance on rating and categorizing these questions.
    \item[Step 2: Define evaluation criteria] We define eleven criteria to evaluate mathematics questions, reported in Table~\ref{tab:criterion}. We construct criteria scales by a Likert scale.
    \item[Step 3: Model evaluation] We input the mathematics questions into ChatGPT and then ask ChatGPT to rate and categorize each question based on the predefined criteria. Finally, we collect ChatGPT's ratings and categorization for all questions.
    \item[Step 4: Result analysis] We compare ChatGPT’s ratings with human annotations, where possible using correlation analysis. We also (1) conduct topic analysis to understand how ChatGPT handles different mathematical topics; (2) evaluate the distribution of questions across various mathematical topics; and (3) assess whether ChatGPT can distinguish between simpler and more complex mathematical problems effectively.
    \item[Step 5: Identify limitations and provide insights] We identify any patterns in ChatGPT’s categorization or rating errors and document any biases or gaps in how the model handles certain topics or difficulty levels. Moreover, we provide insights into the model’s strengths and weaknesses in rating and categorizing mathematics questions and Discuss potential improvements or modifications to enhance ChatGPT’s effectiveness in educational contexts.
\end{description}

\begin{table*}[t]
\caption{Our proposed criterion and metrics for using ChatGPT to rate CSAT Mathematics questions}
\centering
\begin{tabular}{ p{1cm} p{3cm} p{5cm} p{5cm}}
    \toprule
    \textbf{Index} & \textbf{Criteria}  & \textbf{Description} & \textbf{Metrics} \\
    \midrule
    1 & \textbf{Difficulty Level} & Considerations include a number of steps required, abstraction level, and use of advanced concepts.& 1 - Very Easy, 2 - Easy, 3 - Moderate, 4 - Difficult, 5 - Very Difficult. 
    \\
    \midrule
    2 & \textbf{Cognitive Demand}  & Assesses the cognitive skill required (aligned with Bloom's Taxonomy).& 1 - Recall, 2 - Understand, 3 - Apply, 4 - Analyze, 5 - Evaluate/Create. 
     \\
    \midrule
    3 & \textbf{Relevance to Learning Objectives} & Checks alignment with key learning objectives and curriculum standards.& 1 - Irrelevant, 2 - Slightly Relevant, 3 - Moderately Relevant, 4 - Highly Relevant, 5 - Essential.\\
    \midrule
    4 & \textbf{Time Requirement} & Evaluates the expected time required to complete, relative to total test time.& 1 - Very Short, 2 - Short, 3 - Moderate, 4 - Long, 5 - Very Long. \\
    \midrule
    5 & \textbf{Clarity and Ambiguity} & Rates the clarity and ease of understanding the question, avoiding ambiguity.& 1 - Very Unclear, 2 - Unclear, 3 - Neutral, 4 - Clear, 5 - Very Clear.\\
    \midrule
    6 & \textbf{Concept Coverage} & Measures the range of concepts tested within the question.&  1 - Single Concept, 2 - Few Concepts, 3 - Moderate Coverage, 4 - Broad Coverage, 5 - Very Broad Coverage.\\
    \midrule
    7 & \textbf{Scoring Fairness} & Assesses how objectively the question can be graded.& 1 - Highly Subjective, 2 - Some Subjectivity, 3 - Neutral, 4 - Mostly Objective, 5 - Fully Objective.\\
    \midrule
    8 & \textbf{Originality and Engagement} & Evaluates novelty, real-world relevance, or problem-solving appeal of the question.& 1 - Very Commonplace, 2 - Low Engagement, 3 - Moderate Engagement, 4 - High Engagement, 5 - Very Original/Engaging.\\
    \midrule
    9 & \textbf{Discrimination Power} & Measures the question’s ability to distinguish between various proficiency levels.& 1 - Very Low, 2 - Low, 3 - Moderate, 4 - High, 5 - Very High.\\
    \midrule
    10 & \textbf{Error-Prone Nature}  & Rates the likelihood that students might misunderstand or make common errors unintentionally.& 1 - Very Error-Prone, 2 - Somewhat Error-Prone, 3 - Neutral, 4 - Low Error Likelihood, 5 - Minimal Errors Likely.\\
    \midrule
    11 & \textbf{Analytical Depth}  & Evaluate the level of reasoning, proofs, or logic required to solve the question.& 1 - Very Basic, 2 - Basic, 3 - Moderate, 4 - Complex, 5 - Very Complex.\\
\bottomrule
\end{tabular}
\label{tab:criterion}
\end{table*}

\section{Results}
\subsection{Accuracy of ChatGPT-enabled solutions}
\begin{figure}[t]
    \centering
    \includegraphics[width=0.47\linewidth]{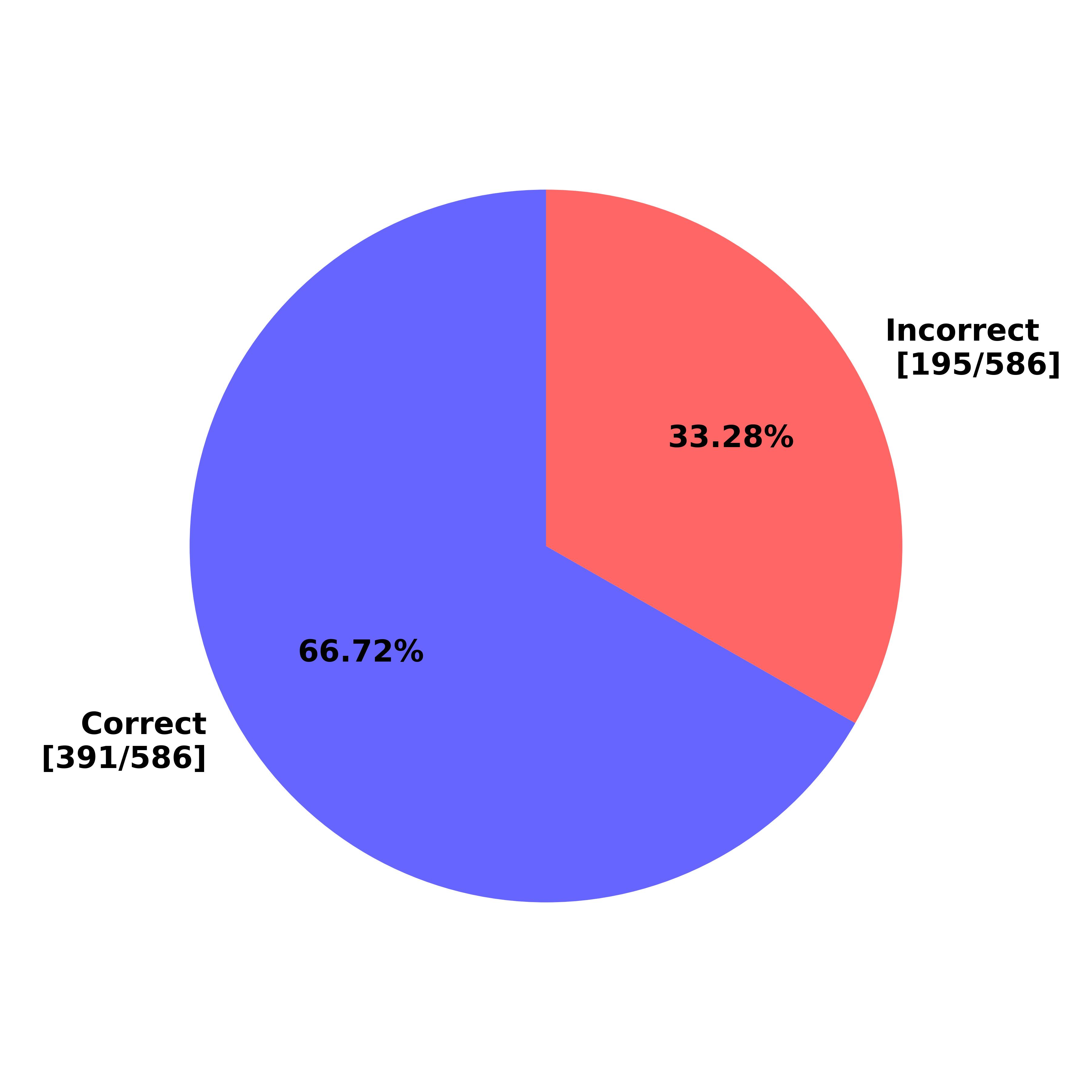}
    \includegraphics[width=0.47\linewidth]{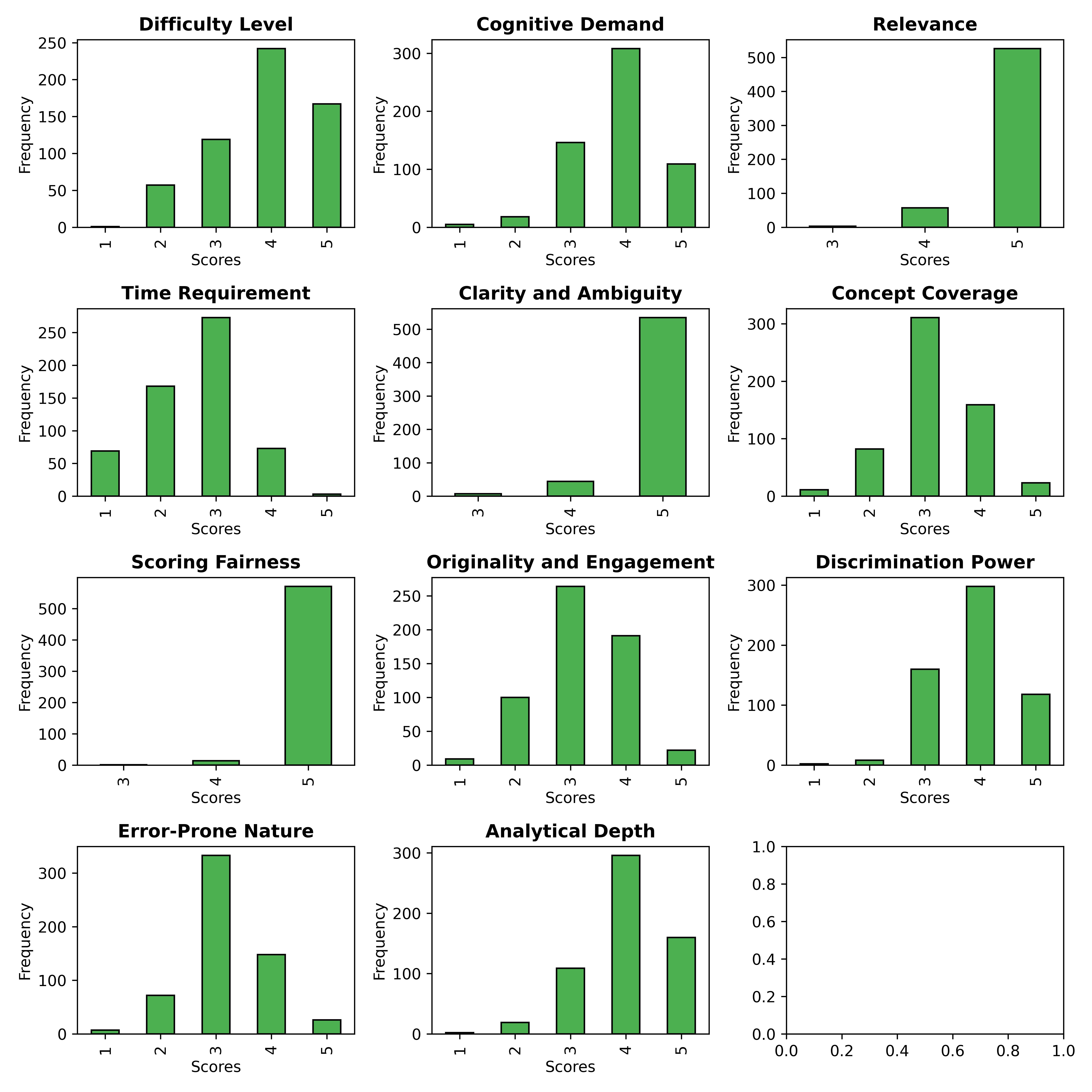}
    \caption{\textbf{(Left)} The accuracy of ChatGPT's solutions for CSAT mathematics questions. \textbf{(Right)} The distribution of score per criteria proposed in Table~\ref{tab:criterion}}
    \label{fig:result_acc}
\end{figure}
Figure~\ref{fig:result_acc} reports the analysis of ChatGPT's performance in tackling CSAT Mathematics questions reveals an overall accuracy rate of 66.72\%, based on a dataset of 586 questions (n = 391 correct responses). This assessment uncovers distinct patterns that illustrate the model's strengths and weaknesses when applied to Korean mathematics problems. Three prominent trends are evident from the findings:
\begin{enumerate}
    \item \textbf{Challenges with sequential questions}: The model often encounters significant difficulties when addressing sequential problems, particularly when the individual sub-questions are considered in isolation. This approach disrupts its ability to maintain logical coherence across interconnected tasks, leading to consistency in the reasoning process.
    \item \textbf{Difficulties with visual information}: The model's notable limitation is its struggle with questions that require the interpretation of visual aids, such as figures, graphs, or diagrams. This lack of proficiency aligns with observations made in the technical documentation for vision-impaired versions of ChatGPT, which indicate challenges in processing graphical information effectively.
    \item \textbf{Incorrect or incomplete responses}: Occasionally, the model generates answers outside the provided options in multiple-choice questions, rendering these answers incorrect within the context of standardized tests. Furthermore, it may produce empty or incomplete responses for fill-in-the-blank questions, which detracts from its overall reliability.
\end{enumerate}
\begin{remark}\normalfont
    These insights emphasize ChatGPT's potential usefulness and limitations in educational and evaluative contexts. Further refinement is essential to enhance its effectiveness, especially in areas requiring sequential reasoning and visual data interpretation. Such improvements could pave the way for a more robust model application in academic settings.
\end{remark}

\subsection{Effectiveness of ChatGPT's ratings}
ChatGPT's comprehensive evaluation of the CSAT Mathematics questions provides valuable insights into the test's perceived difficulty, cognitive demands, and overall characteristics. The following details are the summary of the AI-enabled rating reported Figure~\ref{fig:result_acc}:
\begin{description}
    \item[Difficulty level]: The assessment indicates that a significant proportion of the questions are rated as difficult (score 4), very difficult (score 5), and moderate (score 3), listed in decreasing order of frequency. This Distribution highlights the challenging nature of the CSAT Mathematics exam, suggesting that students must possess a robust understanding of mathematical concepts and problem-solving skills to succeed.
    \item[Cognitive demand]: The evaluation reveals that most test questions require higher-order cognitive skills, ranging from scores 3, 4 to 5. Notably, around 20\% of the questions require higher-level skills such as evaluation and creative problem-solving. In contrast, a smaller segment of questions requires only basic recall (score 1) or understanding of concepts (score 2). This finding indicates that the exam emphasizes critical thinking and the ability to apply mathematical concepts in diverse situations.
    \item[Content relevance]: ChatGPT recognizes the high relevance of the presented questions to the subject of mathematics. However, this observation may be influenced by the inherent mathematical notations and equations contained within the problems, which are fundamental to the discipline. This relevance ensures that students are tested on the skills and knowledge that are essential for success in the field.
    \item[Time requirement]: The model estimates that most questions can be solved within very short to moderate time frames, which is intriguing given the high ratings for both difficulty and cognitive demand. This suggests that while the questions may be challenging, they are designed in a way that allows for efficient problem-solving within the constraints of the exam.
    \item[Analytical depth]: Almost all questions exhibit a high (score 4) to a very high level (score 5) of analytical depth. This indicates that students are required to engage in substantial reasoning, proof-based thinking, and logical deduction to arrive at correct answers. The depth of analysis necessary signals that students must not only know mathematical procedures but also understand the underlying principles that govern these processes.
\end{description}
\begin{remark}\normalfont
    \textbf{Consistency with human perception}: The results of the evaluation are consistent with the perceptions of educators and students regarding the CSAT Mathematics test, which is widely recognized as a challenging assessment. This alignment underscores the reliability of the evaluation process.
\end{remark}
Moreover, ChatGPT recognizes the investigating exam is designed with clear problem statements, broad and fair content coverage, objective scoring, a balanced mix of engaging and original questions, strong discrimination power, and minimal error-prone elements, ensuring a rigorous yet accessible assessment of students’ mathematical competencies. Specifically, we analyze each of these criterion based on Figure~\ref{fig:result_acc} (left panel):
\begin{description}
    \item[Clarity of problem statements]: The problem statements have been rated as highly clear. This clarity is likely a result of the straightforward nature of mathematical language and the use of universal mathematical notations, which help to overcome potential language barriers and make the problems accessible to a broader audience.
    \item[Content coverage and fairness]: ChatGPT suggests that the content covered by the test is fair and includes a moderate breadth (score 3) of mathematical knowledge. Compared to equivalent examinations in other countries, such as Vietnam and the U.S. (SAT or ACT), the CSAT is noted for its broader content coverage, aligning more closely with Advanced Placement (AP) examinations. This broad coverage ensures that the exam tests a wide range of mathematical skills, providing a more holistic evaluation of students’ competencies.
    \item[Scoring fairness]: The model assesses scoring as fully objective (score 5) for the majority of questions, reflecting the structured format of multiple-choice and fill-in-the-blank problems. These formats inherently lack subjective scoring elements, contributing to a fair and transparent assessment process.
    \item[Originality and engagement]: Approximately 50\% of the questions are considered moderately engaging (score 3), while 34\% are identified as highly engaging (score 4). The evaluation also notes that the ratio of commonplace questions to those deemed highly original and engaging is nearly equal. This balance showcases the test's effective mix of traditional and innovative problem types, which can help maintain student interest and motivation.
    \item[Discrimination power]: ChatGPT acknowledges the high discrimination power of the CSAT Mathematics exam. Most questions are rated as having high (score 4) to very high (score 5) abilities to distinguish between varying proficiency levels among students. This means the questions effectively differentiate between those who have mastered the material and those who may still be struggling.
    \item[Error-prone nature]: More than 50\% of the questions are assessed as having a neutral error-prone nature. This aligns with the overall clarity and relevance of the problem statements, demonstrating that the design of the questions minimizes confusion and facilitates accurate responses from test-takers.
\end{description}
\begin{remark}\normalfont
    These findings collectively highlight the strengths, complexities, and nuances of the CSAT Mathematics exam. They reaffirm its reputation as a rigorous and comprehensive assessment tool that evaluates students’ mathematical capabilities and readiness for further academic challenges.
\end{remark}

\subsection{Statistical significance}
Figure~\ref{fig:chisquared_plot} represents the pairwise significance of relationships between various metrics used to evaluate the CSAT Mathematics exam based on p-values from a Chi-squared test. The strongest relationships (indicated by \( p \approx 0 \)) exist between marks, difficulty level, cognitive demand, time requirement, and originality and engagement. This reflects the intertwined nature of these metrics in characterizing the rigor and structure of questions. All pairwise relationships appear significant (no white or light-blue cells), implying that none of the metrics are independent. This aligns with the idea that exam design integrates multiple interrelated factors, such as balancing difficulty with cognitive demand and engagement. Difficulty level and cognitive demand exhibit robust connections to other metrics, confirming their foundational role in defining a question's characteristics. The weaker relationship of accuracy with engagement and originality highlights potential areas for improvement in AI models, such as adapting to non-standard or creative question formats.

\subsection{Correlation analysis}
\begin{figure*}[htb]
    \centering
    \includegraphics[width=\linewidth]{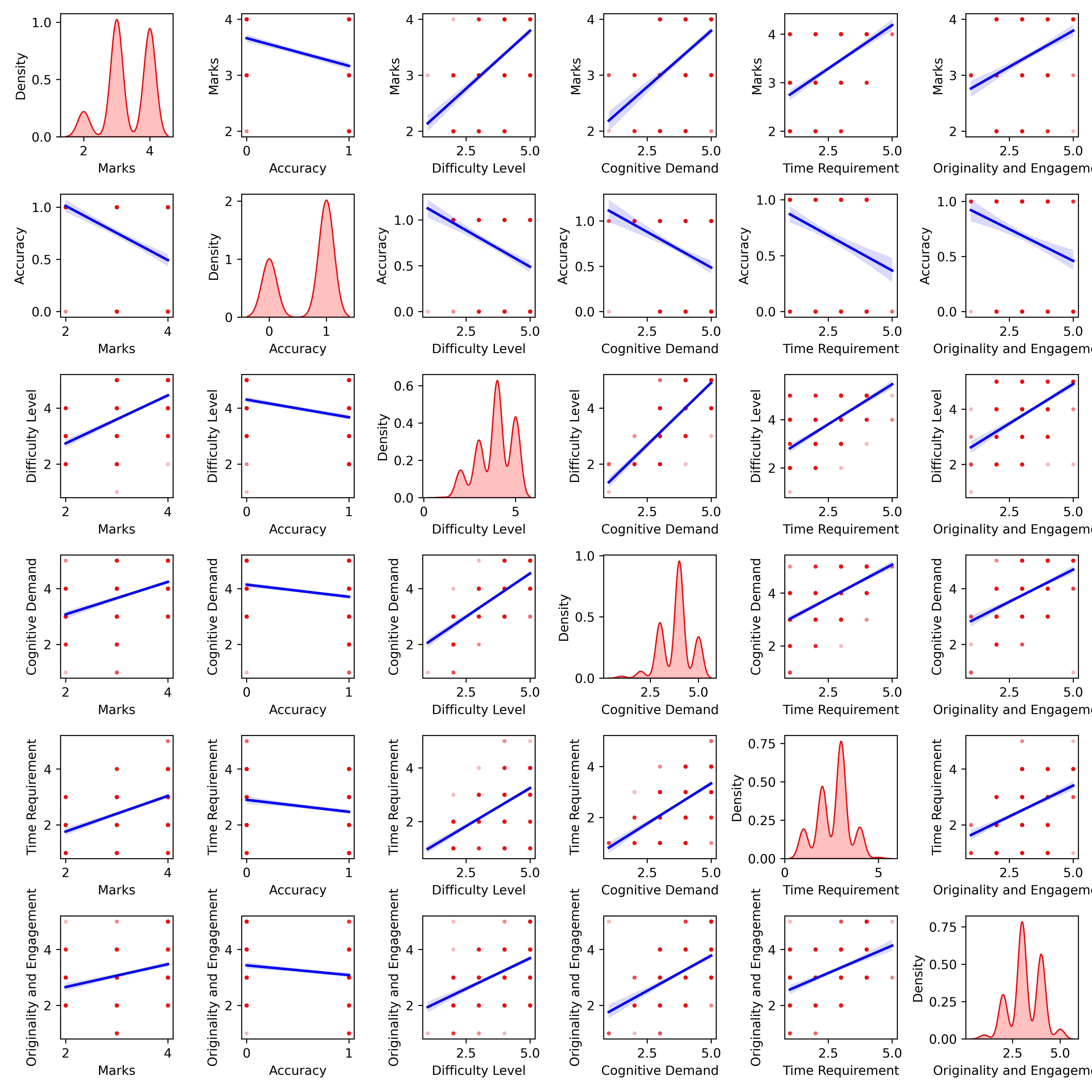}
    \caption{Correlations between actual question marks with: \textbf{(1)} the accuracy of AI solutions, \textbf{(2)} predicted difficulty, \textbf{(3)} predicted cognitive demand, \textbf{(4)} predicted time requirement, and \textbf{(5)} predicted originality and engagement}
    \label{fig:corr_analysis}
\end{figure*}

The diagonal of Figure~\ref{fig:corr_analysis} represents the Distribution of individual variables, typically visualized as histograms or kernel density plots (KDE), whereas off-diagonal plots show pairwise relationships between variables, which are (1) the accuracy of AI solutions, (2) predicted difficulty, (3) predicted cognitive demand, (4) predicted time requirement and (5) predicted originality and engagement. For each scatter plot, we add a linear regression curve to illustrate the correlation between the variables.

Generally, while ChatGPT excels at solving lower-marked questions, its performance declines for more complex, higher-marked problems. Addressing this limitation is essential for improving its capabilities in advanced problem-solving. The AI demonstrates strong alignment in predicting task difficulty, cognitive demand, and time requirements, indicating it can contribute effectively to educational content creation and assessment design. In the following detailed analysis, we will report on the form \textbf{evidence-suggestion}, for which the evidence is based on the illustrated statistical results, followed by our Suggestion to improve ChatGPT for the given problem. 

\subsubsection*{ChatGPT provides reasonable rating}
\begin{description}
    \item[Evidence] \textit{ChatGPT's accuracy in solving problems varies significantly based on the point value of the questions}. For 2-point questions (lower-marked), the accuracy is nearly 100\%, indicating consistent performance on straightforward or less complex problems. In contrast, for 4-point questions (the highest-marked), accuracy drops to approximately 50\%. This decline suggests that ChatGPT needs help with the increased complexity and multi-step reasoning required for higher-marked questions.
    \item[Suggestion] There is a clear need for targeted improvements in the AI's ability to handle advanced problem-solving tasks, particularly in strategies for multi-step reasoning and a deeper understanding of nuanced mathematical concepts.
\end{description}

\begin{description}
    \item[Evidence] \textit{A positive correlation exists between the predicted difficulty level and the actual marks assigned to questions}: ChatGPT consistently predicts that questions with higher point values are more challenging. This indicates that its difficulty estimation aligns well with the test's point-value system.
    \item[Suggestion] This correlation validates ChatGPT's capacity to perceive and model the complexity of questions based on their scoring. Further fine-tuning of this difficulty-prediction mechanism could enhance its performance in educational applications.
\end{description}

\subsubsection*{AI as a potential path towards adaptive learning and personalized education}
\begin{description}
    \item[Evidence] \textit{The predicted cognitive demand shows a positive correlation with the question points}: Higher-marked questions necessitate greater cognitive engagement, as anticipated by ChatGPT. This implies that AI recognizes the need for more advanced cognitive skills (e.g., applying, analyzing, and creating) to solve higher-point questions.
    \item[Suggestion] This correlation highlights ChatGPT's potential to classify tasks based on Bloom's taxonomy levels. This capability can support adaptive learning systems by recommending questions aligned with students’ cognitive skill levels.
\end{description}

\subsubsection*{CSAT Mathematics is a well-designed test}
\begin{description}
    \item[Evidence] \textit{There is a minimal positive correlation between predicted engagement (originality and interest) and question scores:} Engagement levels appear relatively balanced across 2-, 3-, and 4-mark questions. This suggests that questions of varying complexity maintain consistent originality and engagement, regardless of their point value.
    \item[Suggestion] This indicates that the test design maintains interest across all question types. ChatGPT can utilize this insight to foster equitable engagement in question generation for assessments.
\end{description}

\subsubsection*{Consistency to educational theory}
\begin{description}
    \item[Evidence] The predicted cognitive demand shows significant correlations with:
    \begin{itemize}
        \item Actual marks: Higher-marked questions require more advanced cognitive skills from the test takers.
        \item Difficulty level: The cognitive demand increases as the difficulty increases.
        \item Time requirement: More cognitively demanding questions also take longer to solve.
    \end{itemize}
    \item[Suggestion] These correlations align with educational theory, which posits that higher-order cognitive processes necessitate greater time and effort. ChatGPT's ability to recognize these patterns can aid in designing well-calibrated assessments that balance cognitive demand and difficulty.
\end{description}

\begin{remark}\normalfont
    Implementing these improvements will position ChatGPT as a valuable educational tool that addresses learner needs and curriculum standards. We summarize four strategies to achieve the goal: (1) \textit{Enhance multi-step reasoning}: Improve ChatGPT's ability to solve multi-step and higher-order mathematical problems; (2) \textit{Refine difficulty prediction}: Utilize the observed correlations to refine the difficulty and cognitive demand estimation mechanisms; (3) \textit{Support adaptive learning}: Leverage ChatGPT’s insights to develop adaptive learning systems that recommend tasks tailored to individual learners’ skill levels; and (4) \textit{Balance engagement}: Maintain consistent engagement levels across question types to ensure sustained learner motivation.
\end{remark}

\begin{figure*}
    \centering
    \includegraphics[width=\linewidth]{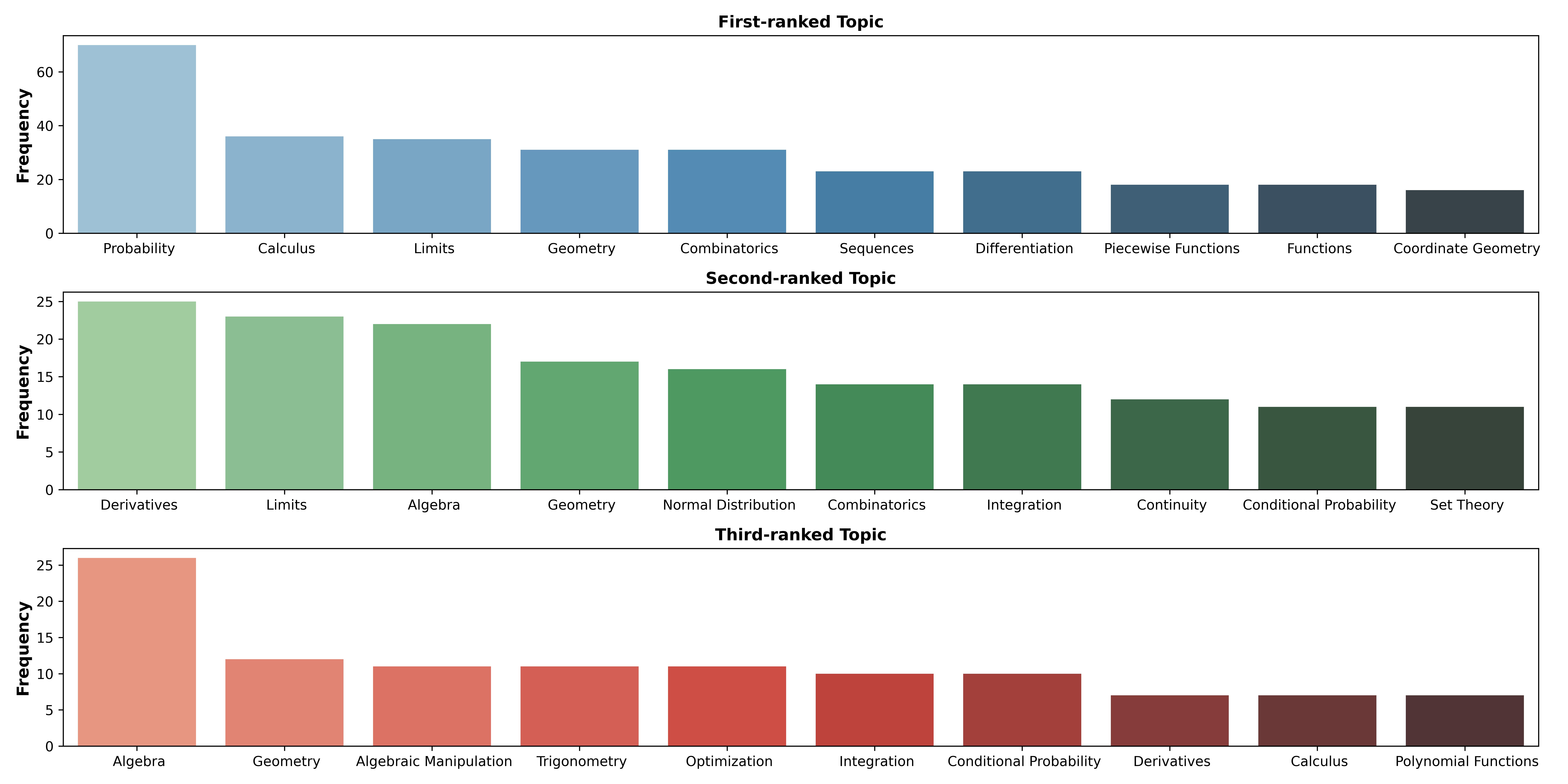}
    \caption{The topic analysis by ChatGPT. We give a detailed discussion in Section~\ref{sec:topic_analysis}}
    \label{fig:topic_distribution}
\end{figure*}
\subsection{Topic analysis}\label{sec:topic_analysis}
In this section, we analyze the content distribution of CSAT mathematics questions, which is reported in Figure~\ref{fig:topic_distribution}. Here, we define the first-ranked topic by the first element in the predicted array from ChatGPT. Similarly, the second-ranked and the third-ranked topics are in the second and third indices. Each question is associated with at least one and at most six topics. Besides, not all questions are categorized with six topics. Specifically, $583$, $479$, $190$, $54$, and $7$ questions are associated with two, three, four, five, and six topics, respectively. We will highlight the mathematical keywords for topics in \texttt{script}.

The most frequent topic in CSAT mathematics is \texttt{probability}, which accounted for roughly $12\%$ ($n=70/568$) of the total questions, followed by \texttt{calculus} and \texttt{limits} with about $6\% (n=35)$ of these collected questions. There are minimal differences in the number of \texttt{geometry} and \texttt{combinatorics} compared to the second-ranked topics, which has $n=31/586$ questions. Regarding the second-ranked topics, \texttt{derivatives} and \texttt{limits} are the most frequent topics, with $n=25$ and $n=23$ questions, respectively. Here, if we aggregate both first- and second-ranked categories, the total number of questions related to \texttt{limits} is $n = 58$. Besides, by assuming \texttt{derivatives} and \texttt{limits} are subjects of \texttt{calculus}, we can conclude that the CSAT mathematics test is mostly focusing on \texttt{probability} and \texttt{calculus}, with lateral priorities concentrated on \texttt{geometry} and \texttt{algebra}.

\begin{figure*}
    \centering
    \begin{subfigure}{0.47\linewidth}
        \includegraphics[width=\linewidth]{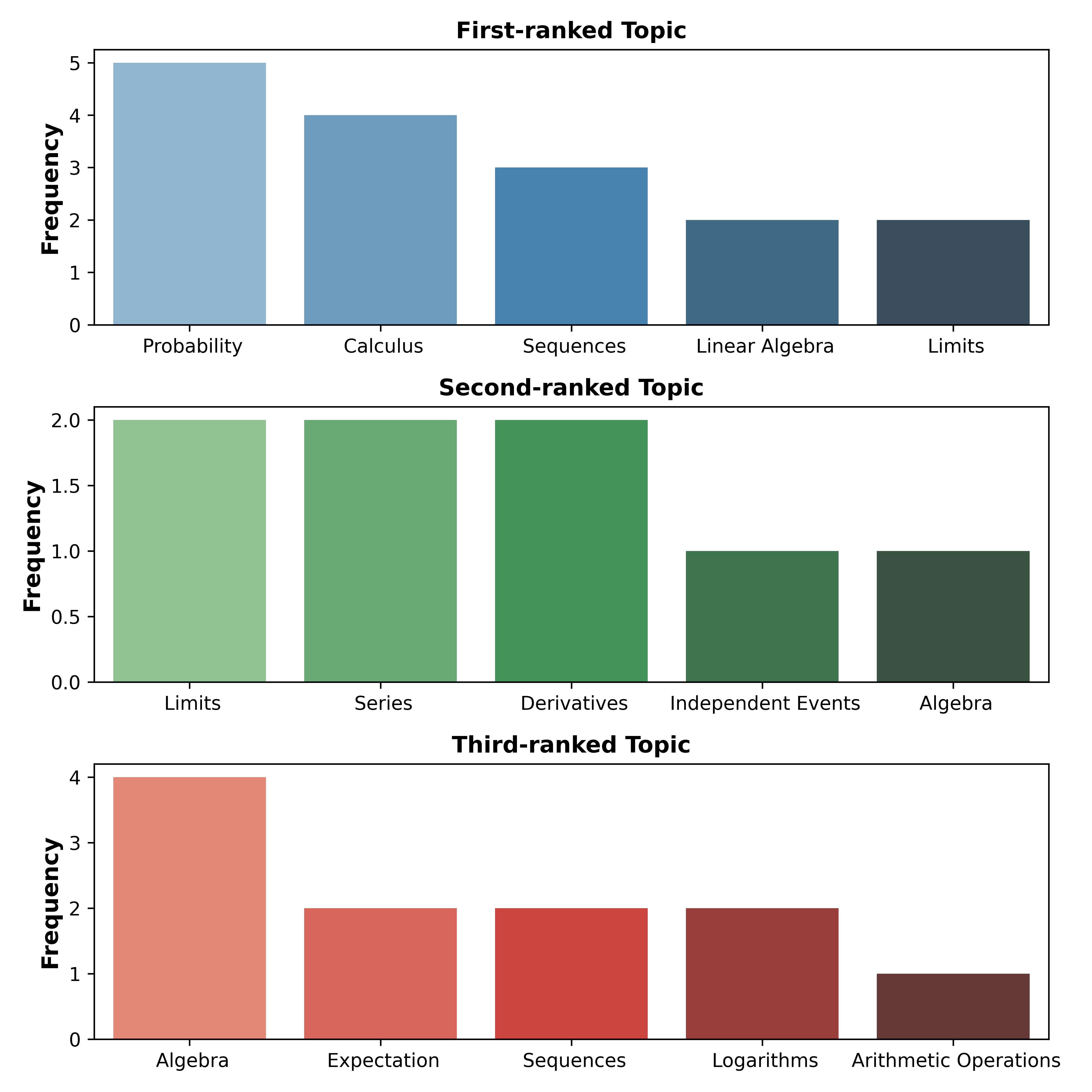}
        \caption{13-11-08-A}
    \end{subfigure}
    \begin{subfigure}{0.47\linewidth}
        \includegraphics[width=\linewidth]{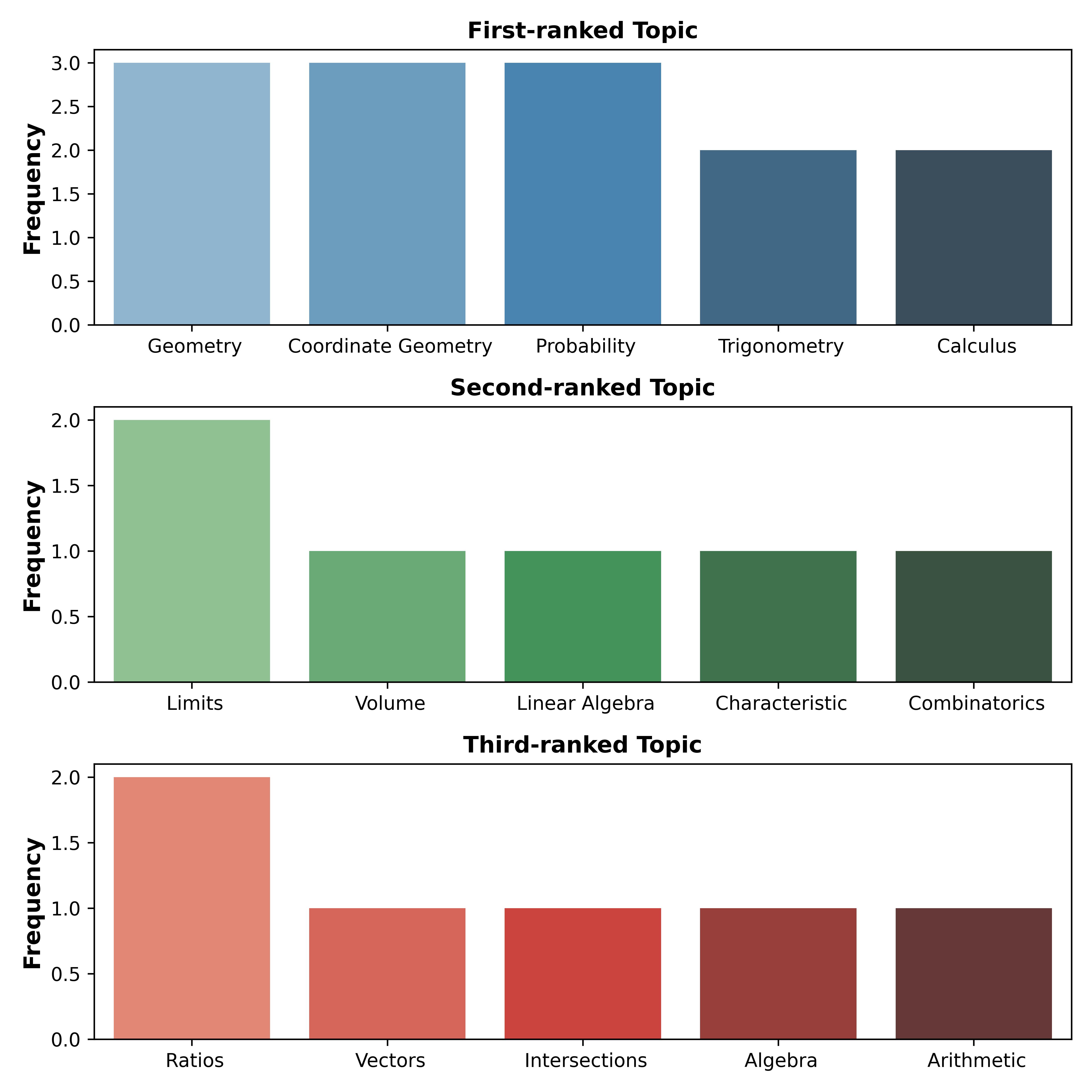}
        \caption{13-11-08-B}
    \end{subfigure}
    \begin{subfigure}{0.47\linewidth}
        \includegraphics[width=\linewidth]{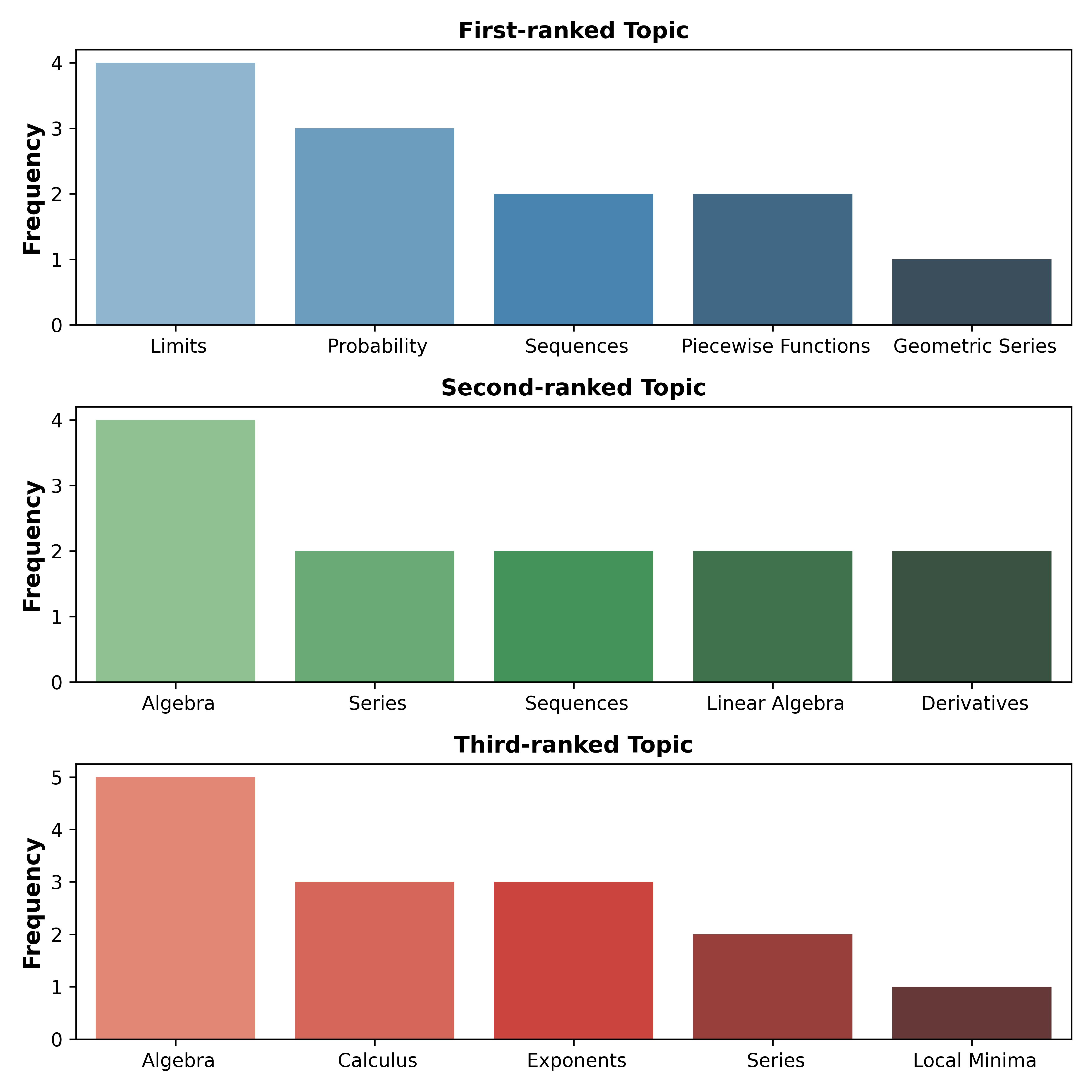}
        \caption{14-11-15-A}
    \end{subfigure}
    \begin{subfigure}{0.47\linewidth}
        \includegraphics[width=\linewidth]{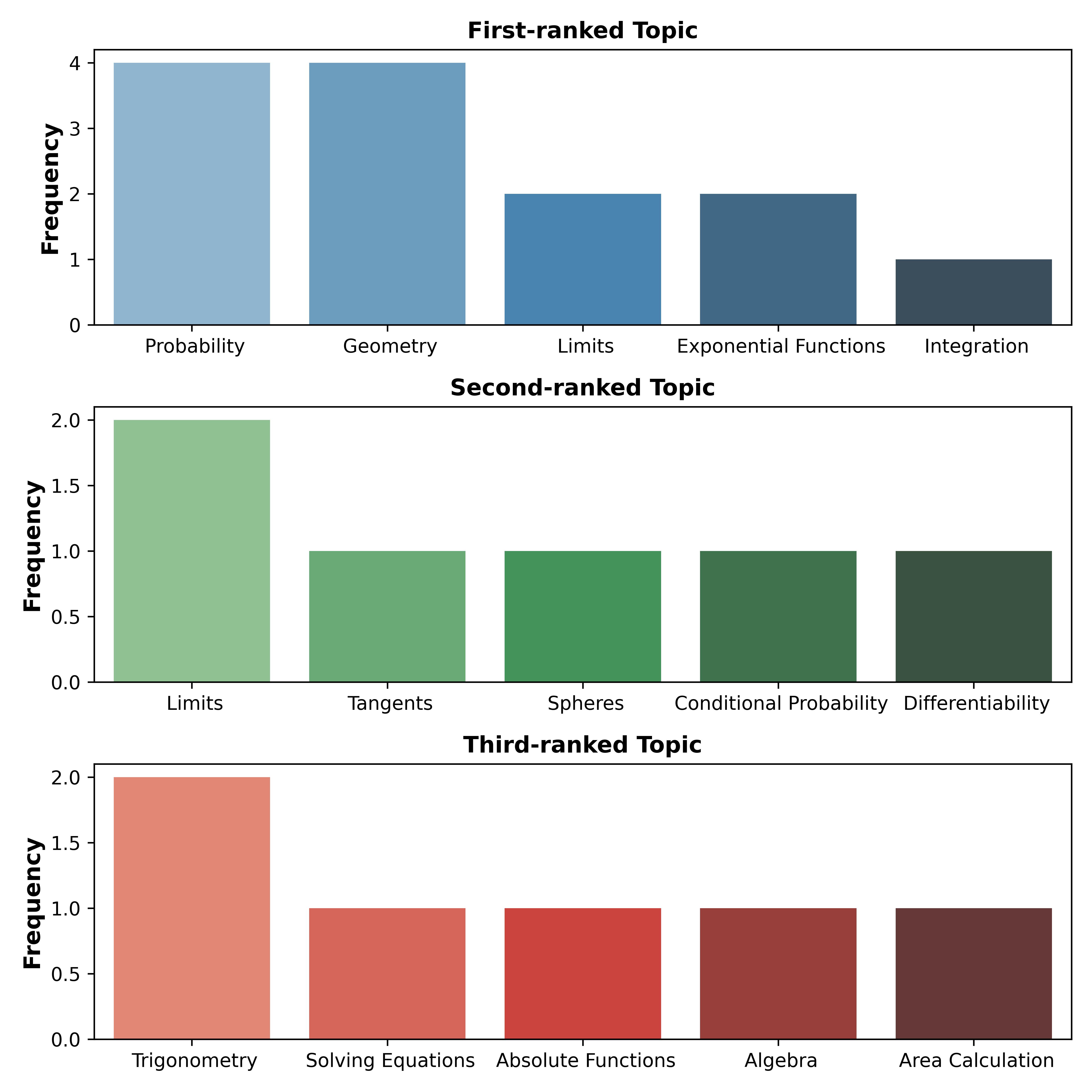}
        \caption{14-11-15-B}
    \end{subfigure}
    \caption{Distribution of topics by test paper (part 1/6)}
    \label{fig:first-set}
\end{figure*}

\begin{figure*}
    \centering
    \begin{subfigure}{0.47\linewidth}
        \includegraphics[width=\linewidth]{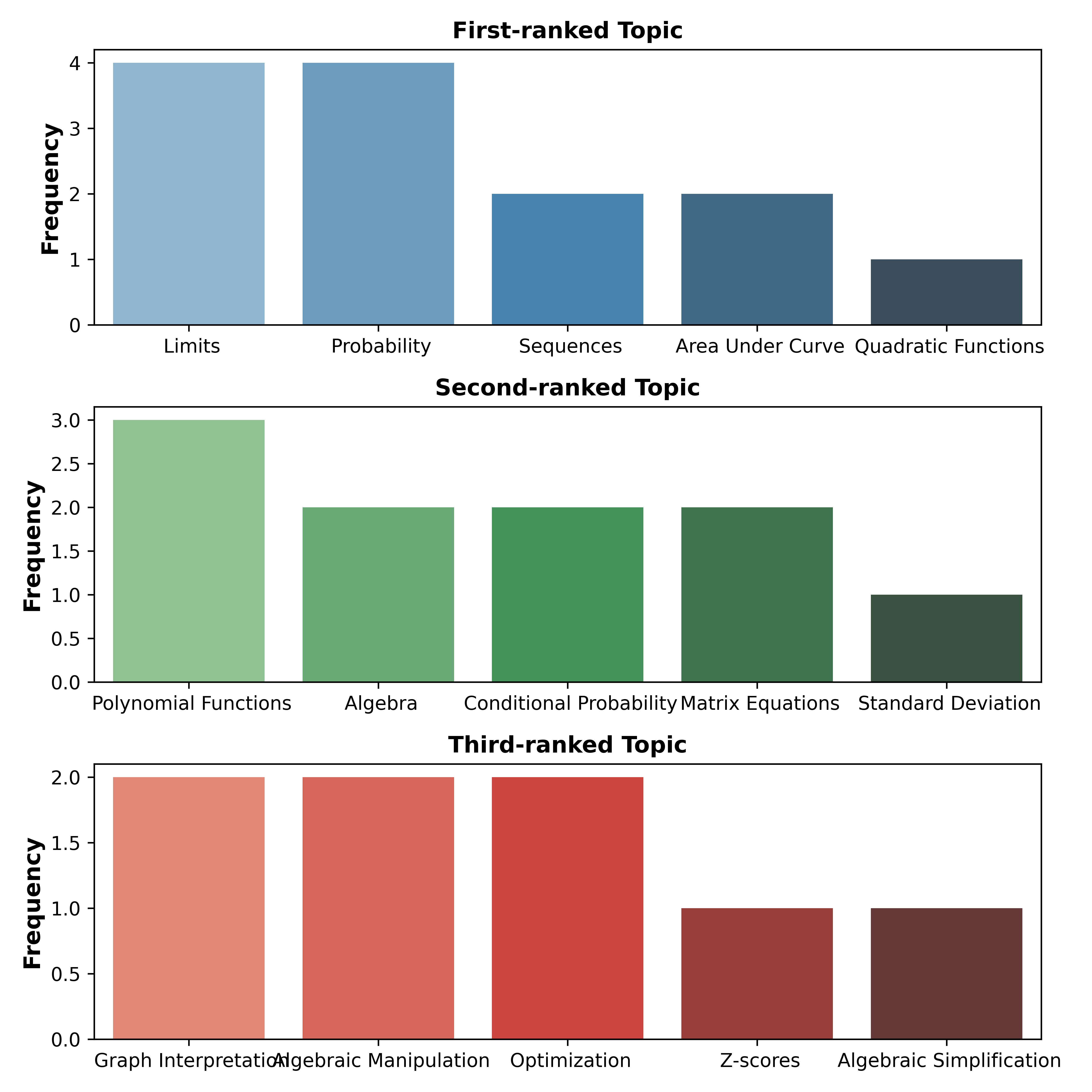}
        \caption{15-11-13-A}
    \end{subfigure}
    \begin{subfigure}{0.47\linewidth}
        \includegraphics[width=\linewidth]{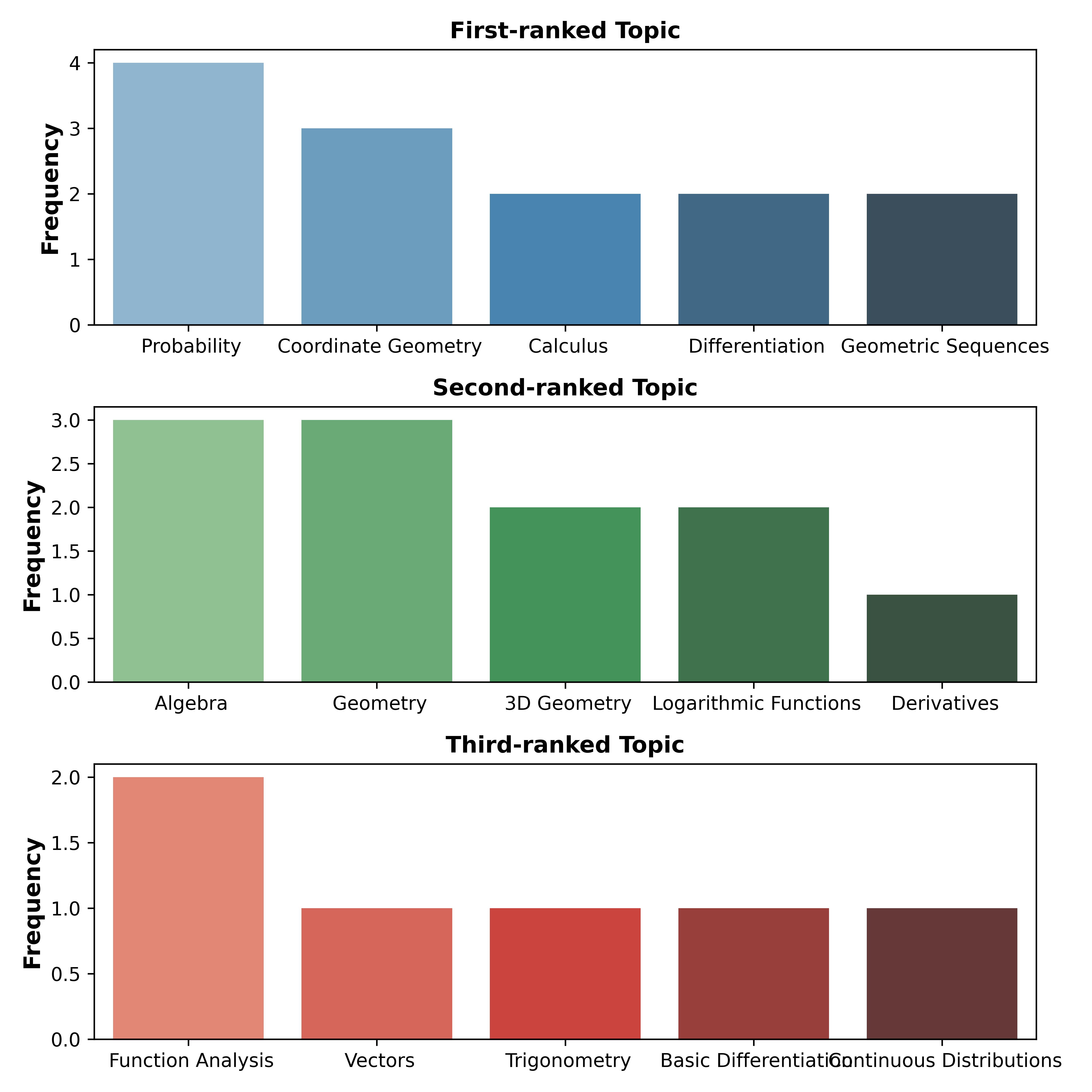}
        \caption{15-11-13-B}
    \end{subfigure}
    \begin{subfigure}{0.47\linewidth}
        \includegraphics[width=\linewidth]{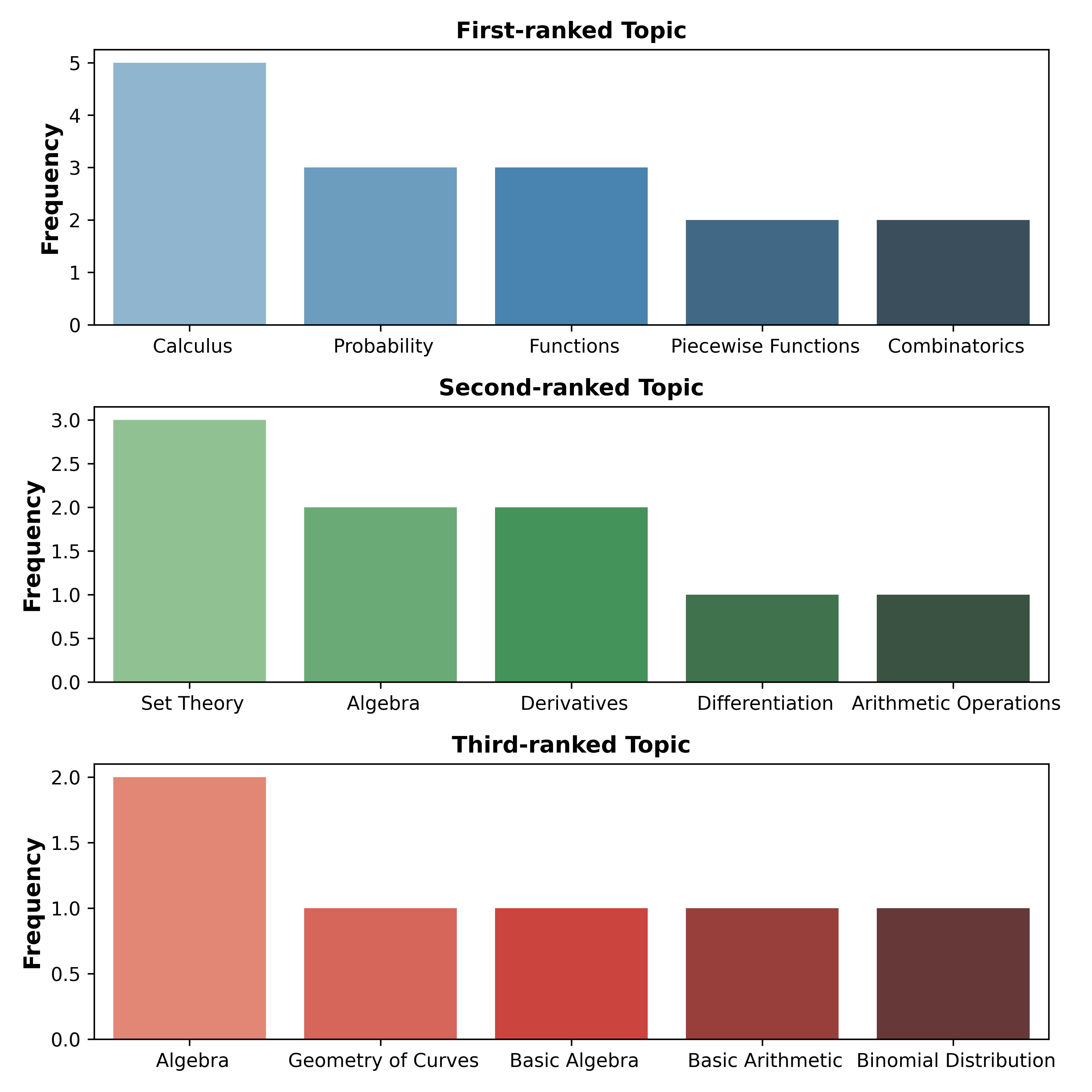}
        \caption{16-11-19-A}
    \end{subfigure}
    \begin{subfigure}{0.47\linewidth}
        \includegraphics[width=\linewidth]{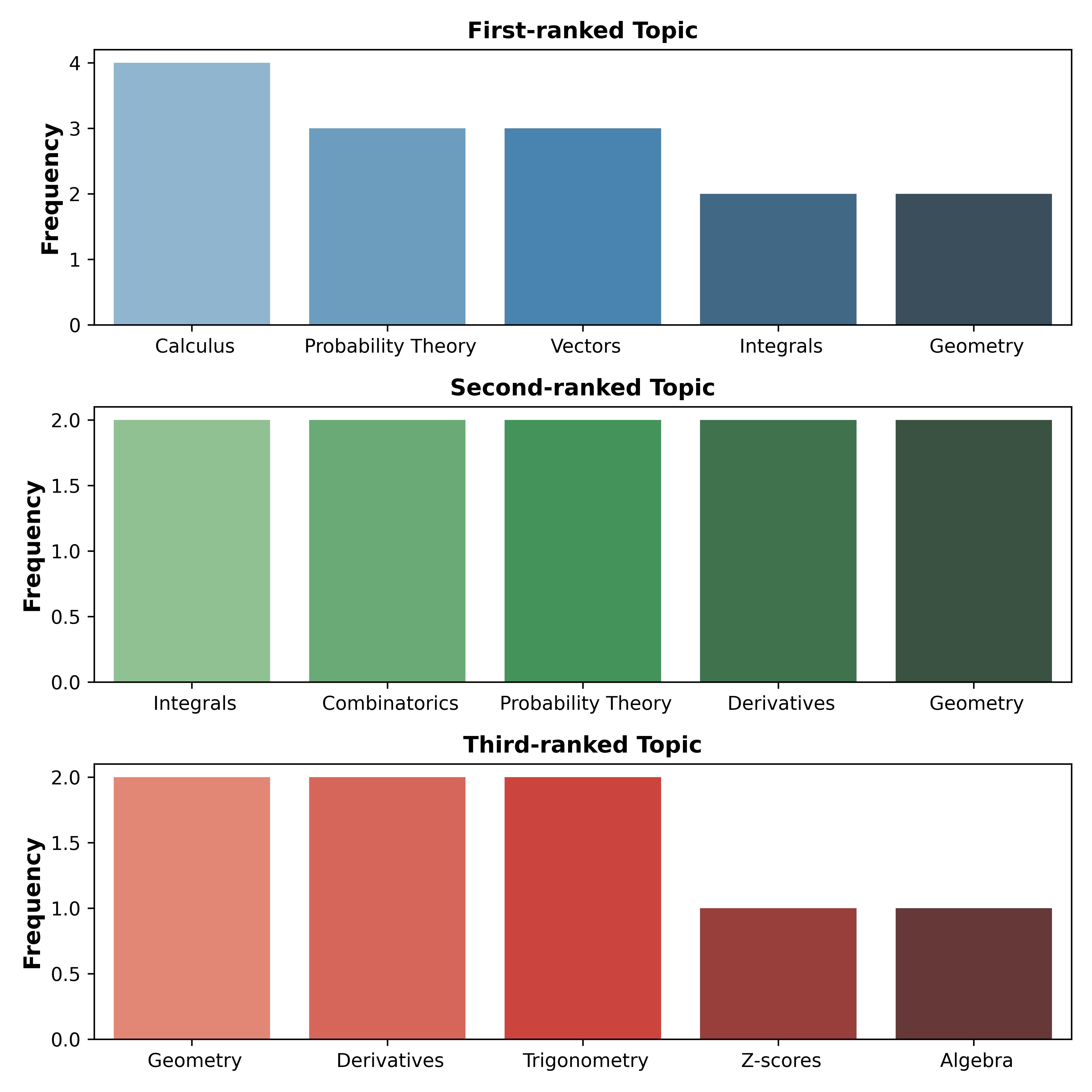}
        \caption{16-11-19-B}
    \end{subfigure}
    \caption{Distribution of topics by test paper (part 2/6)}
    \label{fig:second-set}
\end{figure*}

\begin{figure*}
    \centering
    \begin{subfigure}{0.47\linewidth}
        \includegraphics[width=\linewidth]{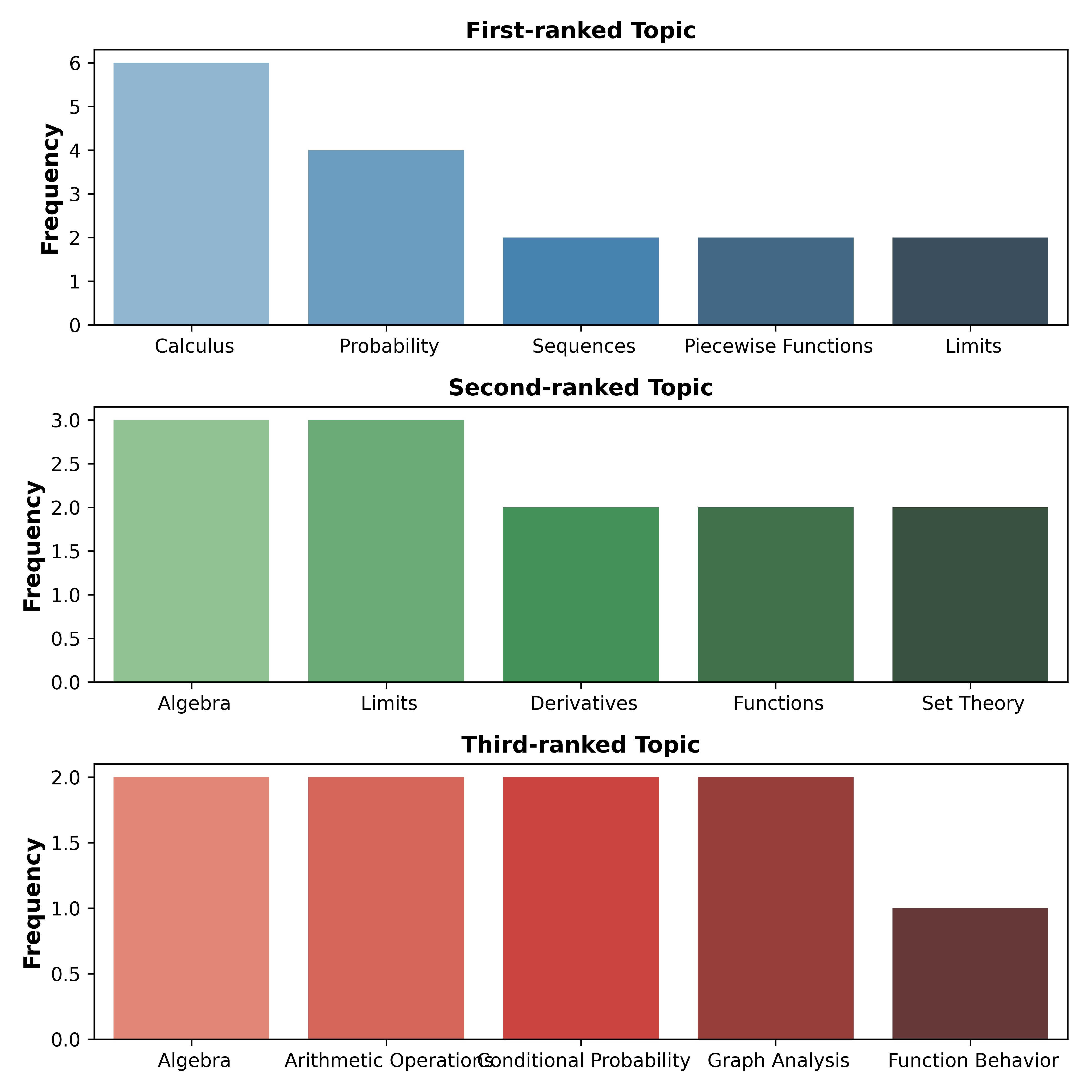}
        \caption{17-11-24-A}
    \end{subfigure}
    \begin{subfigure}{0.47\linewidth}
        \includegraphics[width=\linewidth]{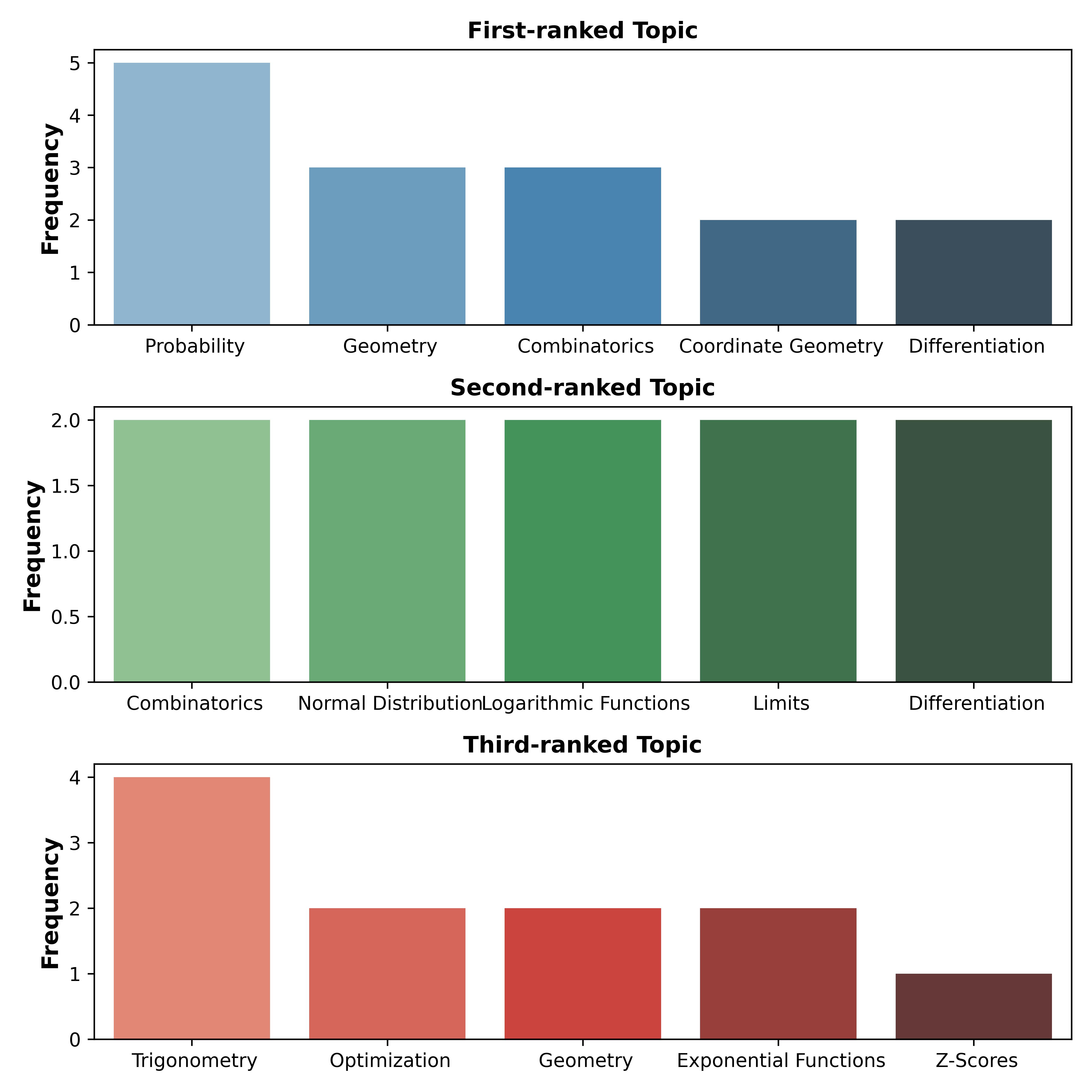}
        \caption{17-11-24-B}
    \end{subfigure}
    \begin{subfigure}{0.47\linewidth}
        \includegraphics[width=\linewidth]{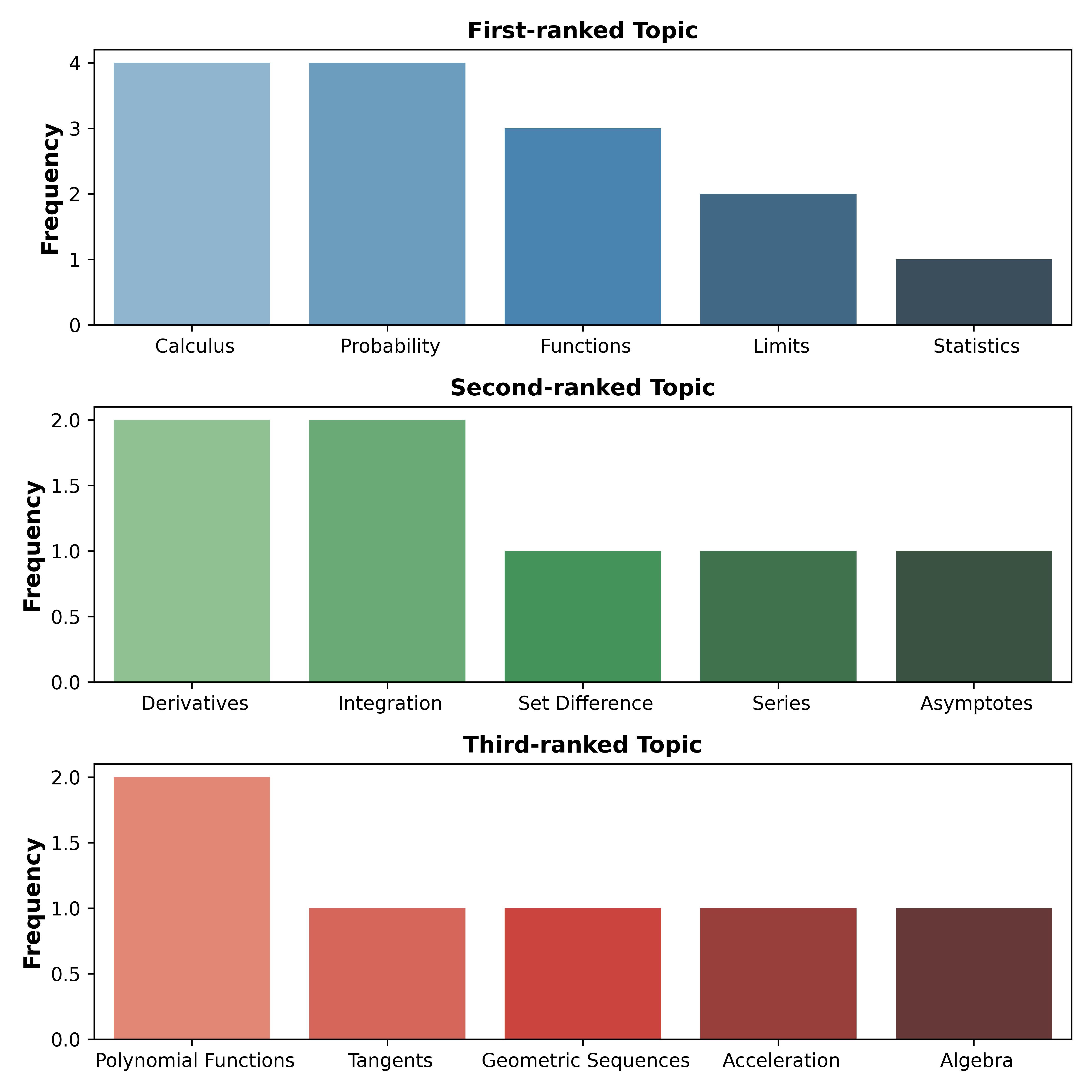}
        \caption{18-11-16-A}
    \end{subfigure}
    \begin{subfigure}{0.47\linewidth}
        \includegraphics[width=\linewidth]{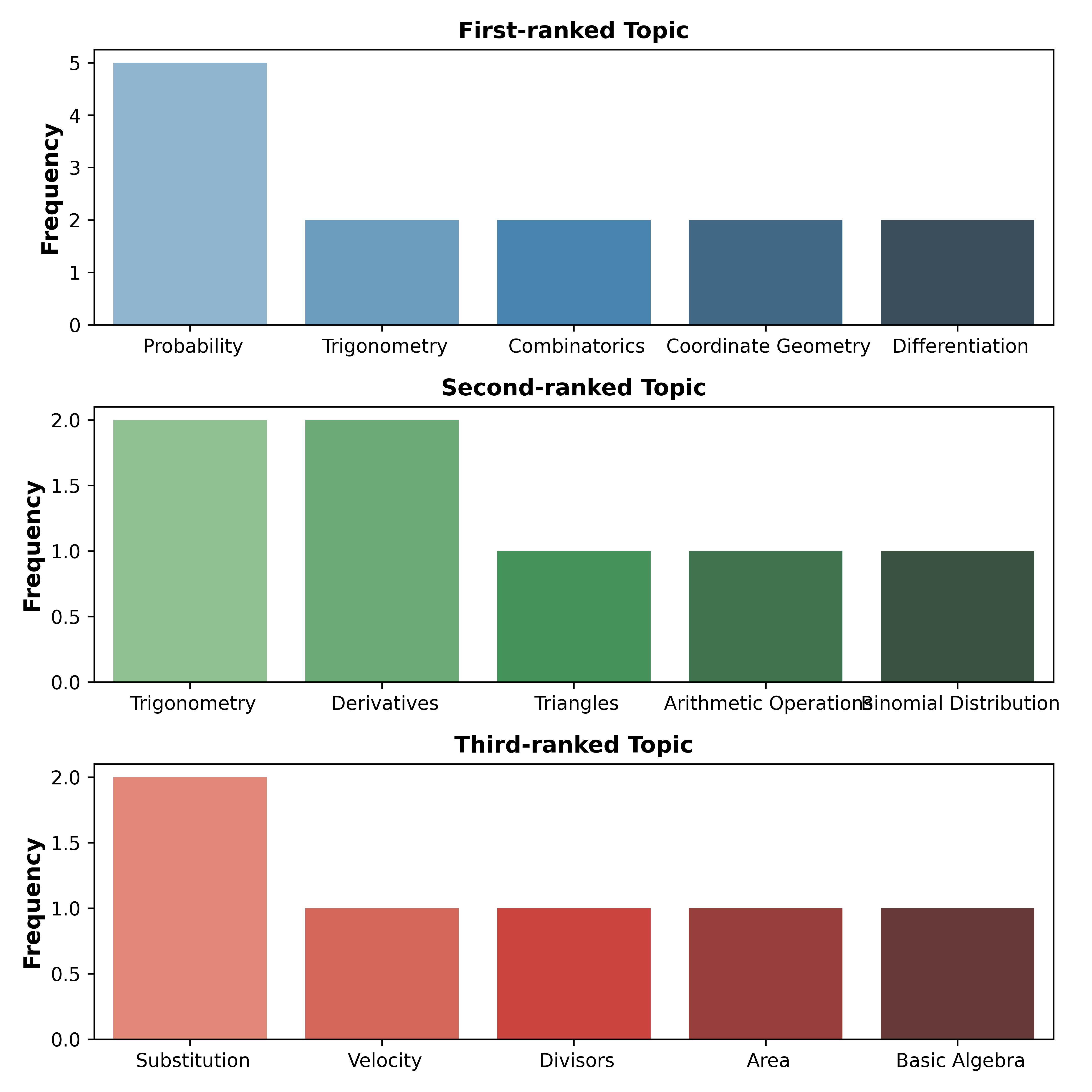}
        \caption{18-11-16-B}
    \end{subfigure}
    \caption{Distribution of topics by test paper (part 3/6)}
    \label{fig:third-set}
\end{figure*}

\begin{figure*}
    \centering
    \begin{subfigure}{0.47\linewidth}
        \includegraphics[width=\linewidth]{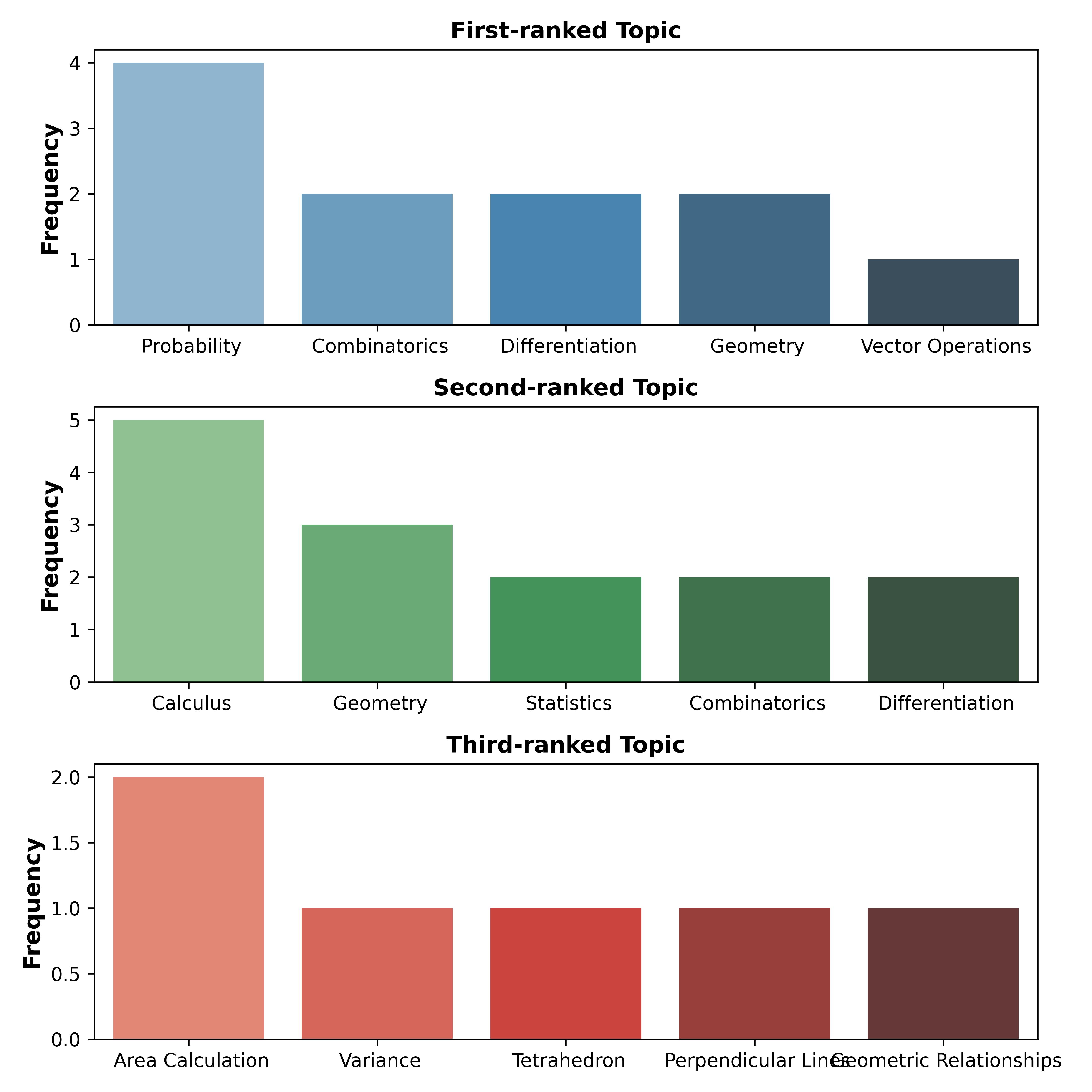}
        \caption{19-11-15-A}
    \end{subfigure}
    \begin{subfigure}{0.47\linewidth}
        \includegraphics[width=\linewidth]{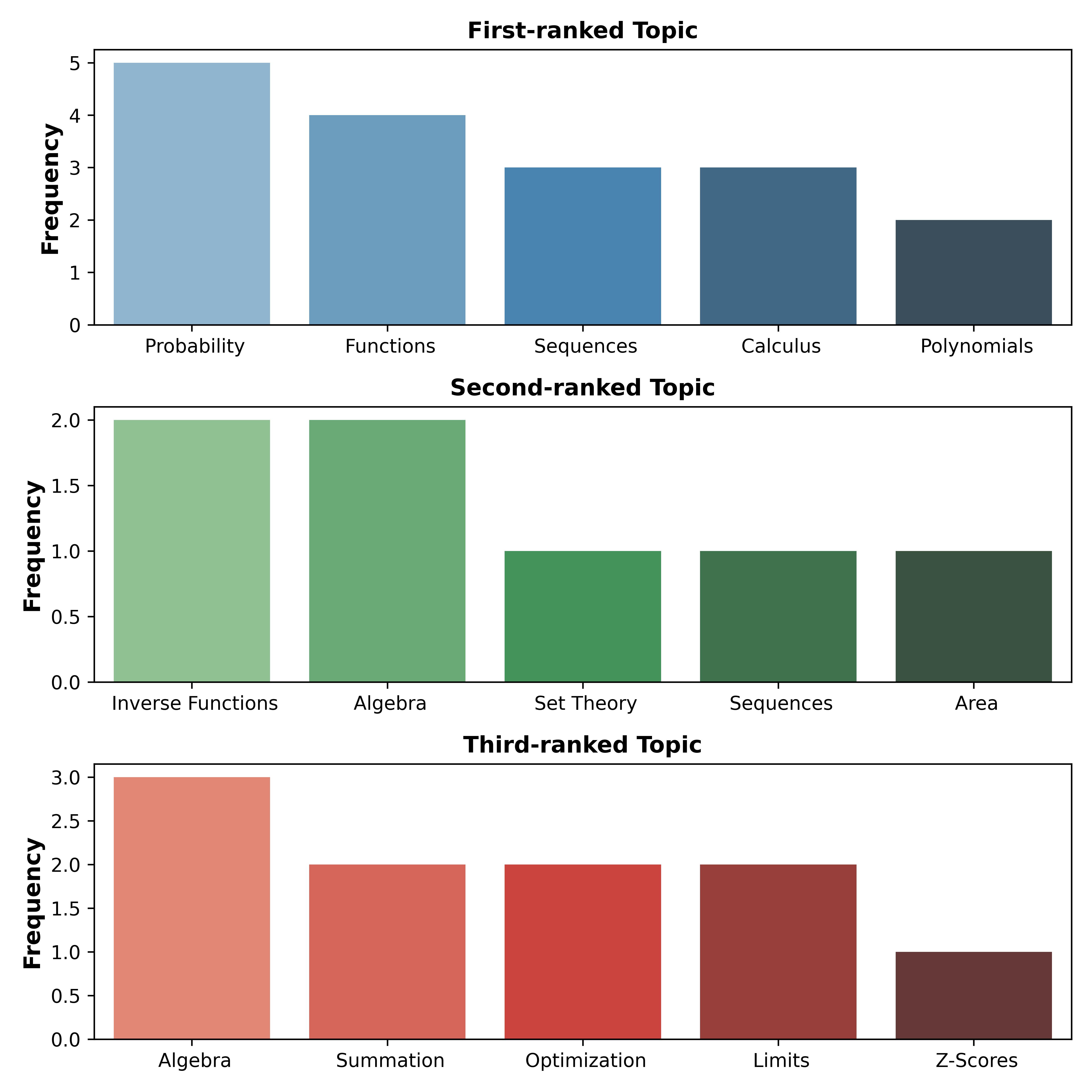}
        \caption{19-11-15-B}
    \end{subfigure}
    \begin{subfigure}{0.47\linewidth}
        \includegraphics[width=\linewidth]{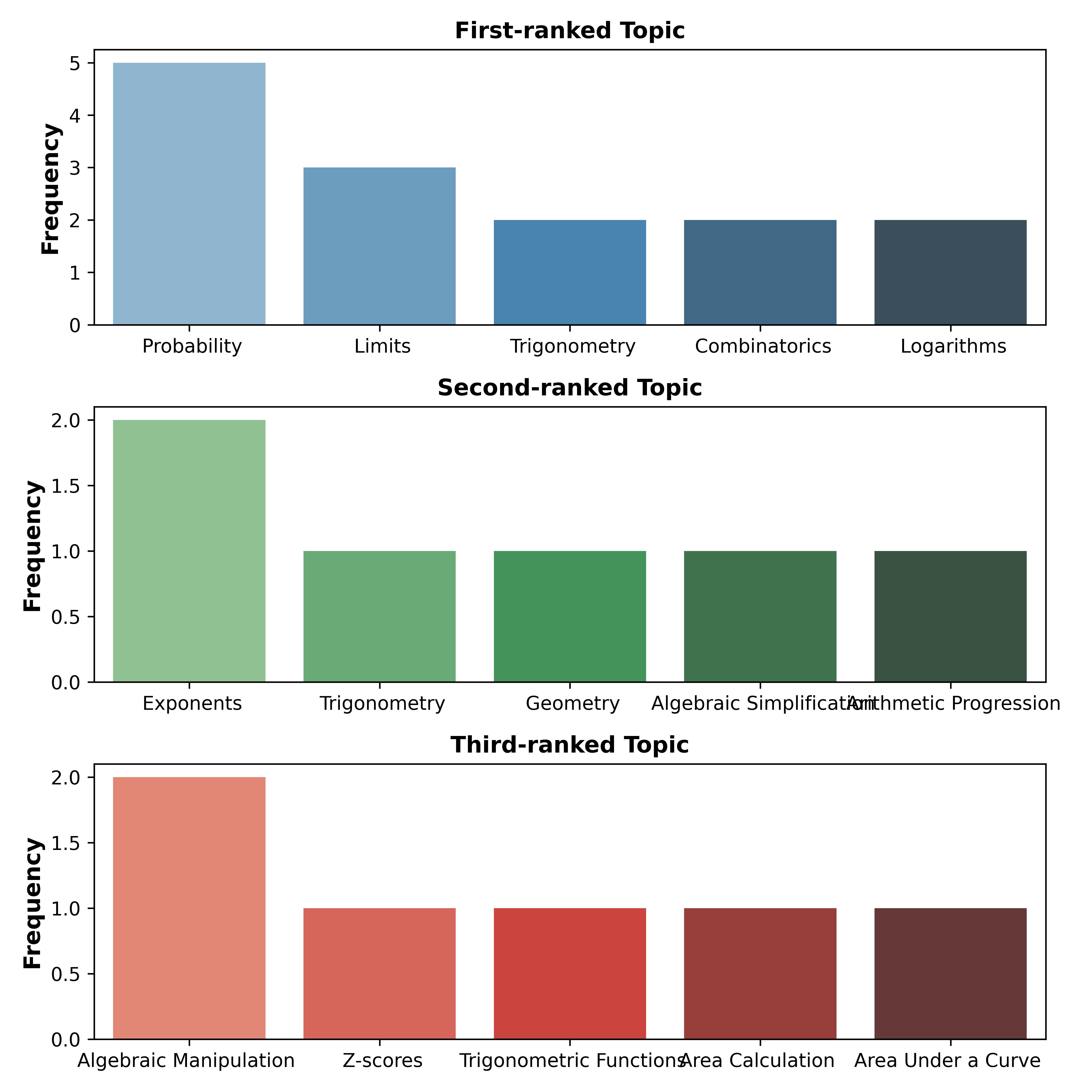}
        \caption{20-12-04-A}
    \end{subfigure}
    \begin{subfigure}{0.47\linewidth}
        \includegraphics[width=\linewidth]{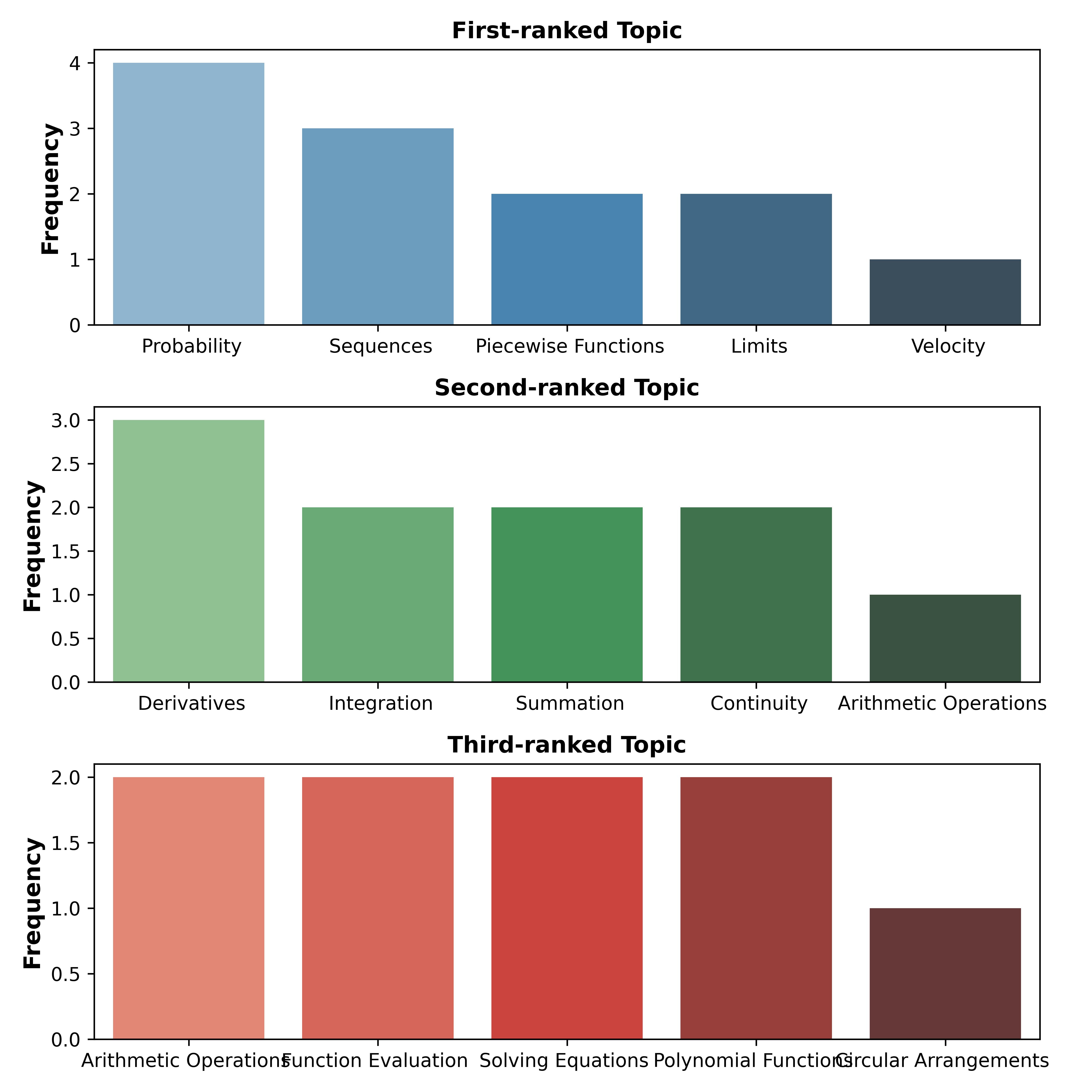}
        \caption{20-12-04-B}
    \end{subfigure}
    \caption{Distribution of topics by test paper (part 4/6)}
    \label{fig:fourth-set}
\end{figure*}

\begin{figure*}
    \centering
    \begin{subfigure}{0.47\linewidth}
        \includegraphics[width=\linewidth]{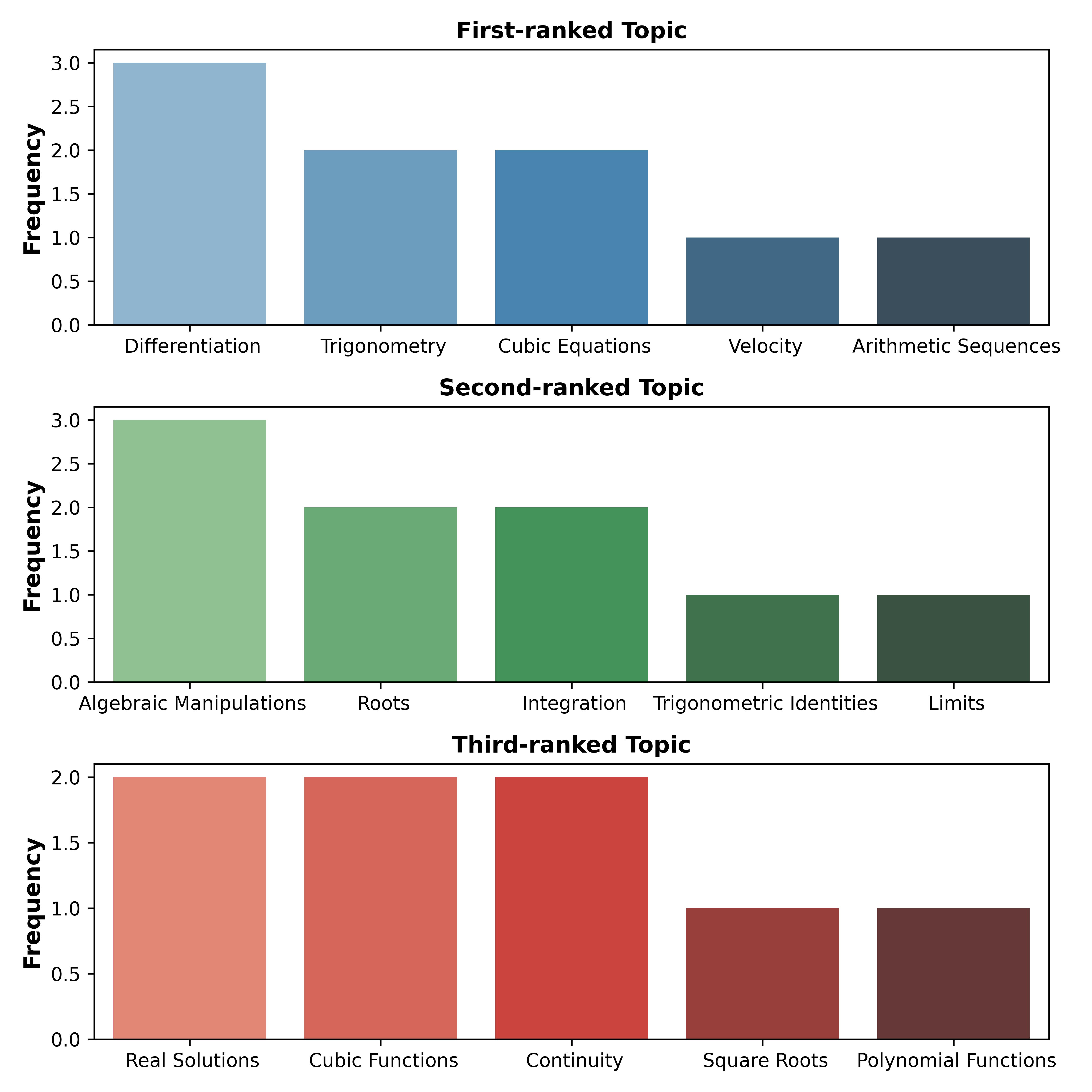}
        \caption{22-11-21-A}
    \end{subfigure}
    \begin{subfigure}{0.47\linewidth}
        \includegraphics[width=\linewidth]{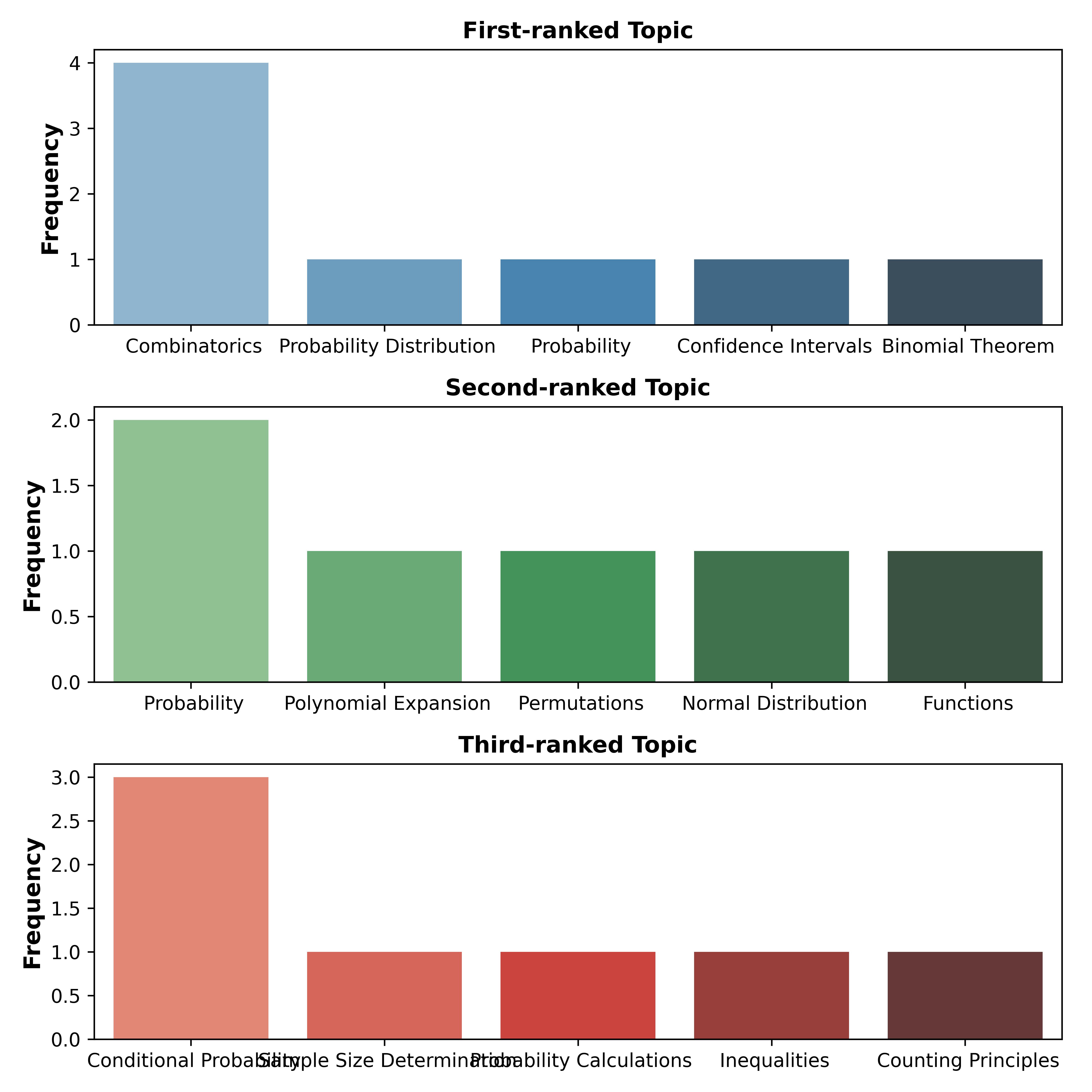}
        \caption{22-11-21-B}
    \end{subfigure}
    \begin{subfigure}{0.47\linewidth}
        \includegraphics[width=\linewidth]{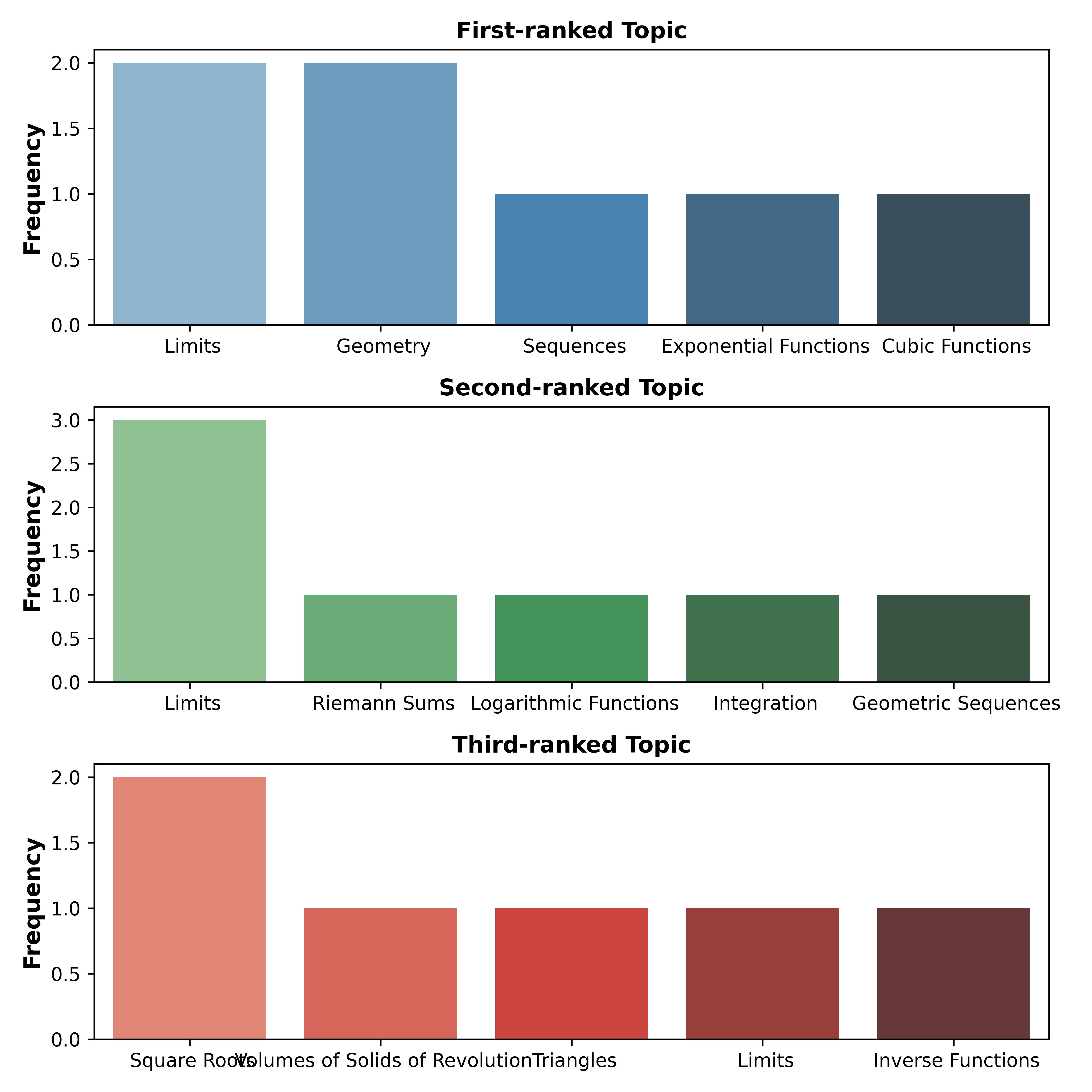}
        \caption{22-11-21-C}
    \end{subfigure}
    \begin{subfigure}{0.47\linewidth}
        \includegraphics[width=\linewidth]{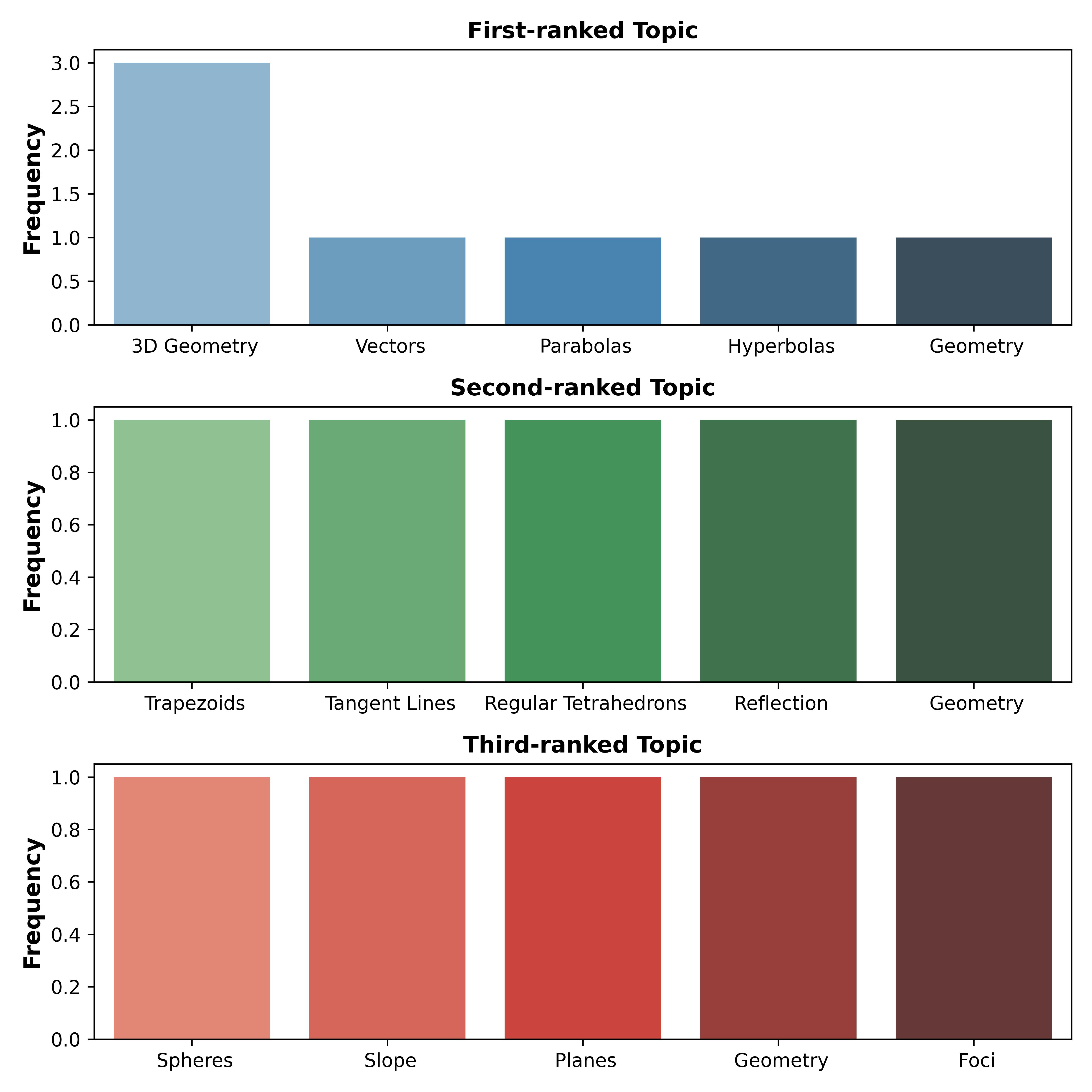}
        \caption{22-11-21-D}
    \end{subfigure}
    \caption{Distribution of topics by test paper (part 5/6)}
    \label{fig:fifth-set}
\end{figure*}

\begin{figure*}
    \centering
    \begin{subfigure}{0.47\linewidth}
        \includegraphics[width=\linewidth]{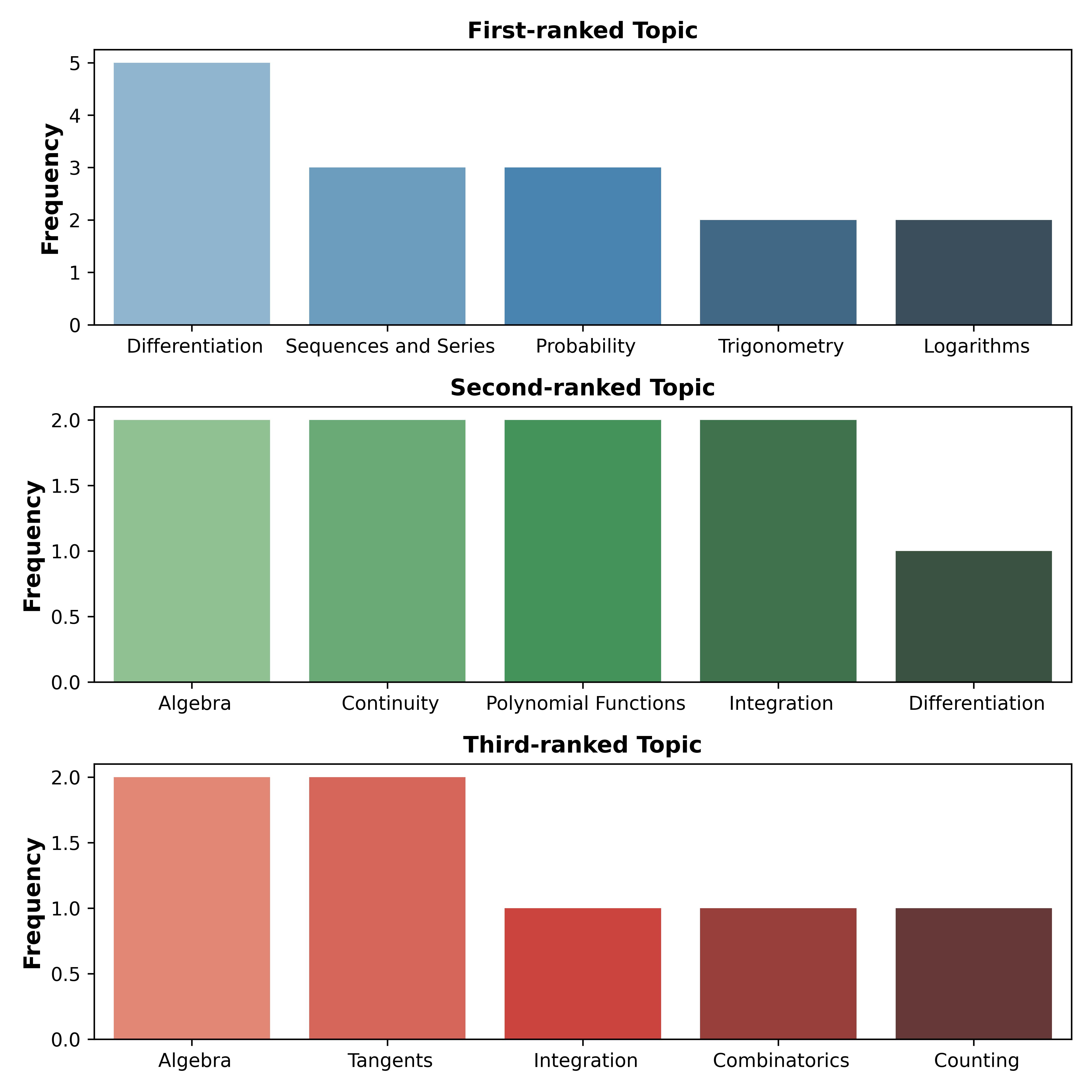}
        \caption{21-11-22-A}
    \end{subfigure}
    \begin{subfigure}{0.47\linewidth}
        \includegraphics[width=\linewidth]{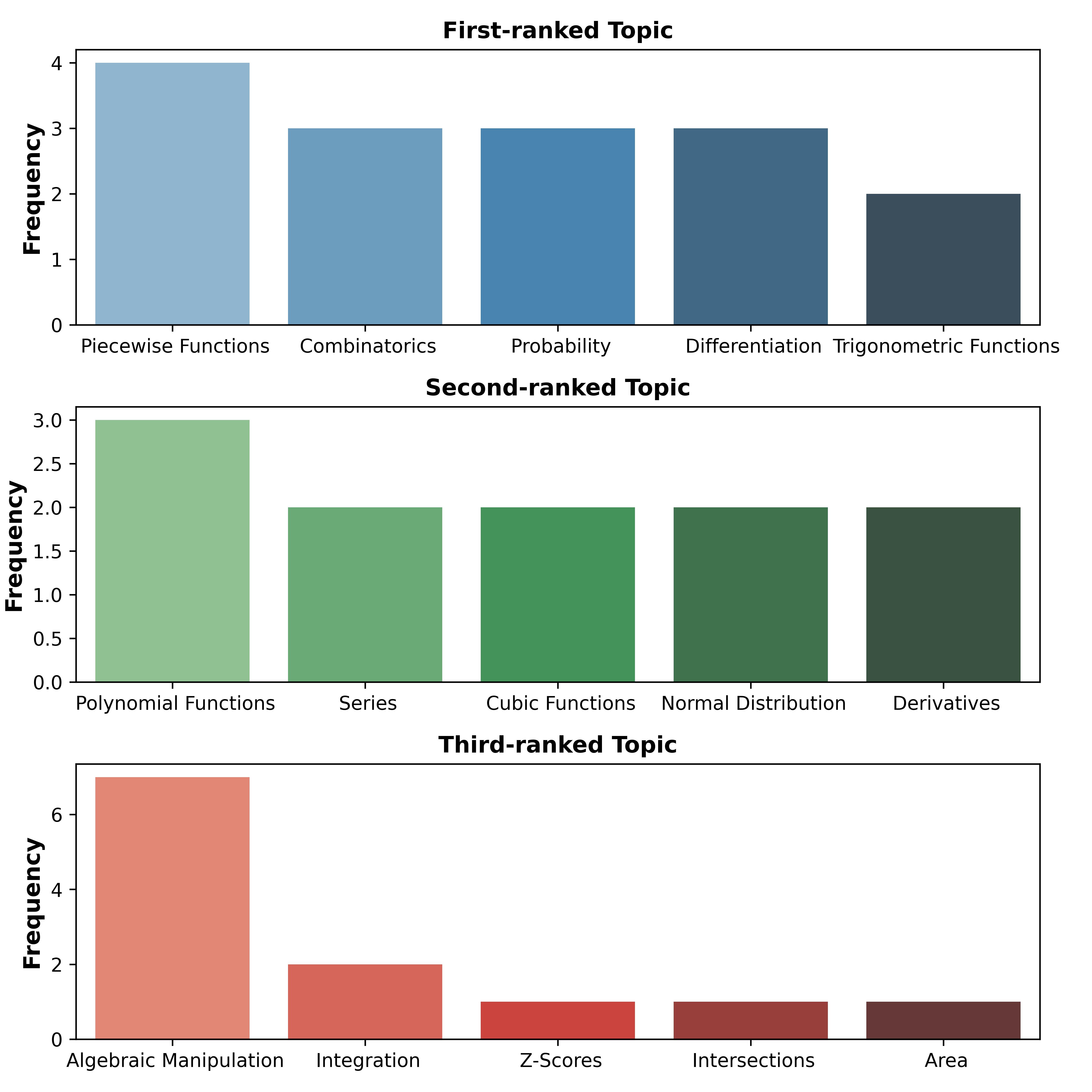}
        \caption{23-11-20-A}
    \end{subfigure}
    \caption{Distribution of topics by test paper (part 6/6)}
    \label{fig:sixth-set}
\end{figure*}

\begin{figure*}
    \centering
    \includegraphics[width=\linewidth]{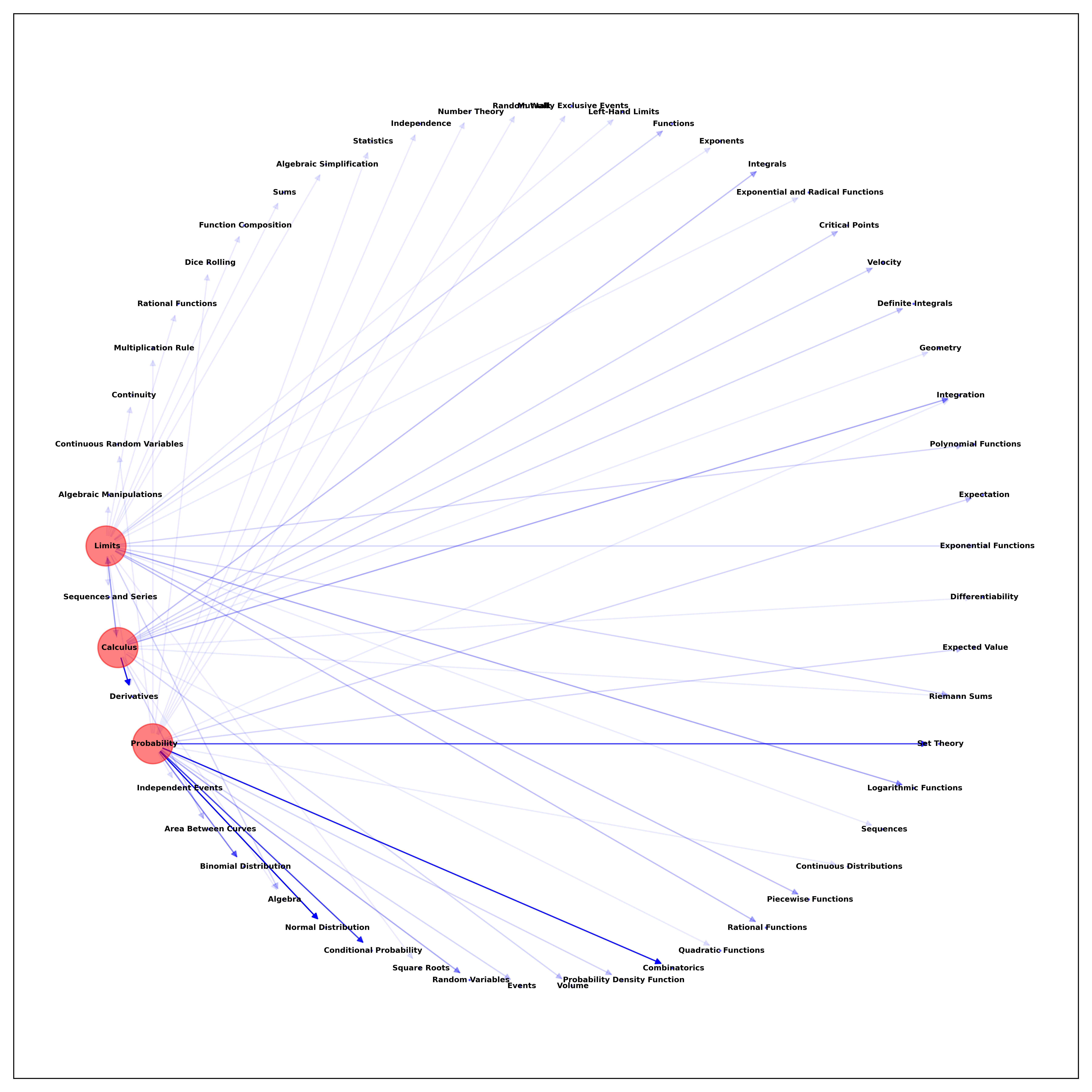}
    \caption{The network of topics in CSAT mathematics questions}
    \label{fig:graph}
\end{figure*}

Using our proposed topic ranking strategy, there is an interesting pattern from ChatGPT inference, observed in Figure~\ref{fig:first-set} to ~\ref{fig:sixth-set}. Particularly, the first- and second-ranked topics are broader topics in mathematics, while the third-ranked topics are specific problem-solving skills. For example, the test \texttt{13-11-08-A} (Figure~\ref{fig:first-set}) has several questions requiring \texttt{arithmetic operations}, which is consistent with both \texttt{probability}, \texttt{limits} and \texttt{algebra}. Another example is test \texttt{14-11-15-B}, which focuses on \texttt{trigonometry}, \texttt{solving equation}, and \texttt{area calculation}. This pattern can be seen in some remaining tests, such as \texttt{15-11-13-A} requires \texttt{graph interpretations} and \texttt{algebraic simplification}; \texttt{16-11-19-A} requires to solve one question about \texttt{binomial distribution} in three \texttt{probability} questions; and so on. From this observation, we not only see that ChatGPT is good at ranking and categorizing CSAT mathematics questions but also good at recognizing related preliminary problem-solving skills. To provide a comprehensive view, we depict the topic connection by a network in Figure~\ref{fig:graph}: three mostly seen topics are in red nodes, which are \texttt{probability}, \texttt{calculus} and \texttt{limits}; the directed edges are highlighted based on the frequency that the first-ranked topics and second-ranked-topics being recognized in the same question.

\section{Discussion}
\subsection{Topics of incorrect responses}
\begin{figure}[h]
    \centering
    \includegraphics[width=\linewidth]{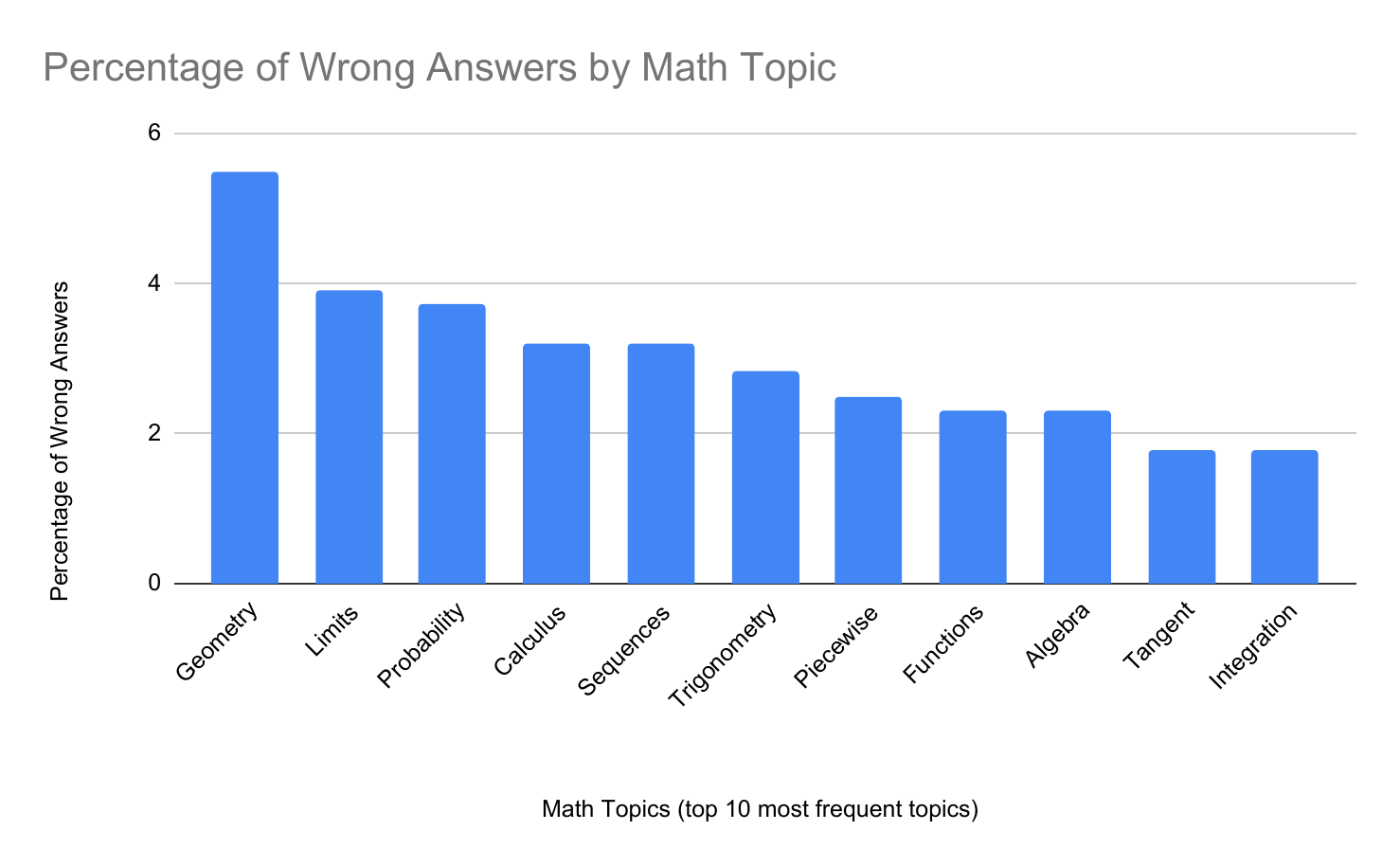}
    \includegraphics[width=\linewidth]{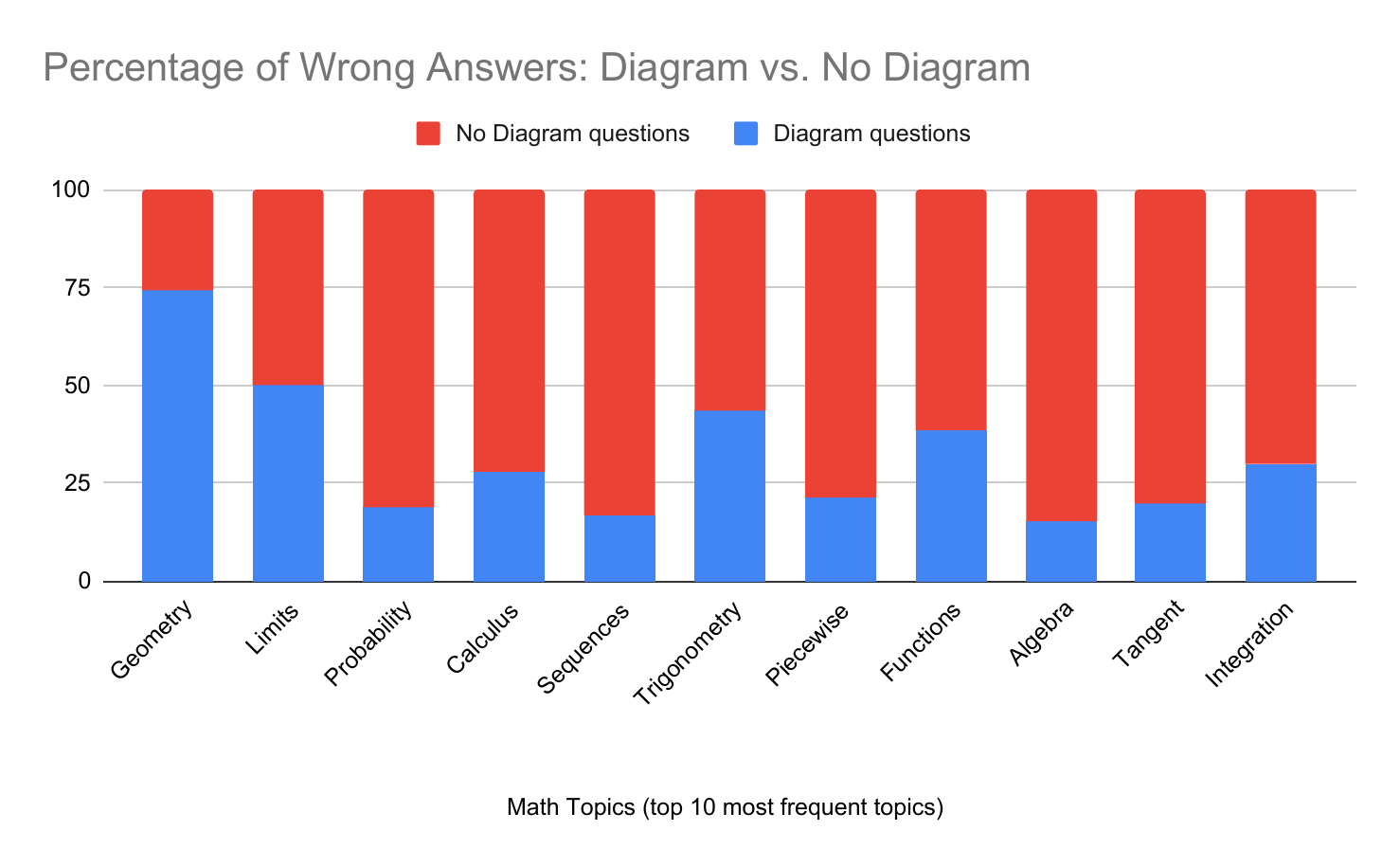}
    \caption{Topics of incorrect responses}d
    \label{fig:topic_wrong}
\end{figure}
Figure~\ref{fig:topic_wrong} (top panel) displays the percentage of incorrect answers for the top 10 most frequent math topics. \texttt{Geometry} and \texttt{Limits} exhibit the highest percentages of incorrect answers, indicating these topics may be particularly challenging for learners. Topics like \texttt{Tangent} and \texttt{Integration} have the lowest percentages of incorrect answers, suggesting they are comparatively well-understood. Topics such as \texttt{Probability}, \texttt{Sequences}, and \texttt{Trigonometry} fall in the mid-range for error rates, representing moderate difficulty.

The lower panel of Figure~\ref{fig:topic_wrong} compares the error rates for questions with and without diagrams across the same 10 math topics. Questions with diagrams have significantly lower error rates compared to those without diagrams, emphasizing the importance of visual aids in this topic. In \texttt{Trigonometry} and \texttt{Piecewise Functions}, a similar trend is observed, where diagram-based questions yield fewer errors, indicating diagrams might help clarify complex visual concepts. These topics \texttt{Algebra} and \texttt{Tangent} show minimal difference in error rates between diagram and non-diagram questions, suggesting that diagrams may not play a critical role in understanding these topics. In most topics, questions with diagrams generally have lower error rates, underscoring the effectiveness of visual aids in improving comprehension and accuracy.

\begin{remark}\normalfont
    Diagram-based questions seem especially beneficial in these challenging areas. Diagrams are instrumental in enhancing understanding, particularly in visually intensive topics like \texttt{Geometry} and \texttt{Trigonometry}. For topics like \texttt{Algebra}, the impact of diagrams is less pronounced, suggesting that textual or symbolic explanations might suffice. For a good CSAT mathematics score, lecturers should prioritize \texttt{Geometry} and \texttt{Limits} in teaching strategies and provide additional support or resources. Besides, practice materials should incorporate diagrams wherever possible, especially in areas where they have shown a significant reduction in error rates. Using a mix of diagram and non-diagram questions can cater to diverse learning styles and better gauge understanding.
\end{remark}

\subsection{Further analysis of incorrect responses}
\begin{table}[t]
    \centering
    \small
    \caption{Model Accuracy and Error Correction Analysis. We further evaluate three prompt engineering techniques to improve the accuracy of ChatGPT in the incorrect-answered questions, including three categories: (Set 1) questions with diagram; (Set 2) questions without diagram; and; all incorrect-answered questions (Incorrect). $\uparrow n$ is denoted the number of correctly answers and $\%$ is the percentage of such questions type in all $586$ questions. There are two key observations: (1) GPT-4o is a more powerful model than GPT-3.5 with consistent improvement in all evaluated sets, and (2) translating Korean questions to English before asking the AI model to solve the questions is more effective.}
    \label{tab:accuracy_analysis}
    \begin{tabular}{l c c c c c c}
        \toprule
        \multirow{2}{*}{\textbf{Strategy}} & 
        \multicolumn{2}{c}{\textbf{(Set 1) Questions with Diagram}} & 
        \multicolumn{2}{c}{\textbf{(Set 2) Non-diagram Questions}} & 
        \multicolumn{2}{c}{\textbf{Incorrect}} \\
        & $\uparrow n$ & \textbf{\%} & $\uparrow n$ & \textbf{\%} & $\uparrow n$ & \textbf{\%} \\
        \midrule
        GPT-3.5 + Korean Q \newline $\rightarrow$ Bilingual A & 
        0 & 48.94 & 0 & 70.12 & 0 & 66.72 \\
        
        GPT-4o + Korean Q \newline $\rightarrow$ Bilingual A & 
        3 & 52.13 & 9 & 71.95 & 12 & 68.77 \\
        
        GPT-4o + English Q \newline $\rightarrow$ Bilingual A & 
        10 & 59.57 & 32 & 76.63 & 42 & 73.89 \\
        
        GPT-4o + English Q \newline $\rightarrow$ English A & 
        7 & 56.38 & 42 & 78.66 & 49 & 75.09 \\
        \bottomrule
    \end{tabular}
\end{table}

\subsubsection*{Addressing low-resource language problems}
Table~\ref{tab:accuracy_analysis} demonstrates a significant improvement in accuracy when prompts are translated into English, underscoring the challenges associated with low-resource languages such as Korean. \textbf{We will denote the number of correctly answered questions by further prompt engineering as $\uparrow n$; i.e., we only consider $195$ questions that ChatGPT provided wrong answer using zero-shot evaluation}. Specifically, while the overall accuracy only marginally increased from 66.72\% ($\uparrow n=0$) to 68.77\% (\(\uparrow n=12\) when comparing GPT-3.5 and GPT-4o models; the accuracy increased significantly to 73.89\% (\(\uparrow n=49\)) when the input questions and prompts have been translated into English beforehand. Moreover, despite having the same input, when we generate English answers only instead of bilingual Korean-English answers, our accuracy increases an additional 1.2\% (from 73.89\% to 75.09\%). Diagram-based questions showed an improvement from 48.94\% (\(\uparrow n=0\)) to 56.38\% (\(\uparrow  n=7\)), while non-diagram questions improved from 70.12\% (\(\uparrow  n=0\)) to 78.66\% (\(\uparrow  n=42\)) using this strategy.
\begin{remark}\normalfont
    Korean is considered a low-to-medium resource language, which has limited training data compared to English. This leads to significant performance gaps. Fine-tuning open-source models on bilingual datasets (Korean-English math problems) or employing direct translation methods, as demonstrated in this study, can help mitigate these challenges
\end{remark}

\subsubsection*{Addressing sequential questions}
\begin{table}[!ht]
    \centering
    \caption{Comparison of Valid Entries and Token between Correct and Incorrect Answers}
    \label{tab:token-usage}
    \begin{tabular}{lcc}
        \toprule
        \textbf{Statistic} & \textbf{Incorrect} & \textbf{Correct} \\
        \midrule
        Total number of questions in $195$ incorrectly answered questions       &  125 & 70 \\
        Average number of steps           & 10.08 & 9.16 \\
        Average completion tokens         & 1009.23   & 902.39 \\
        Average prompt tokens             & 791.70    & 721.51 \\
        Average total tokens              & 1800.93   & 1623.90 \\
        \bottomrule
    \end{tabular}
\end{table}
Among the $n=195$ incorrectly answered questions, $n=125 (21.33\%))$ remained unsolved across any of the tested scenarios. On average, these unsolved questions require more steps to solve ($10.08$ steps vs. $9.16$ steps for questions solved successfully by at least one strategy). They also feature longer problem descriptions and solutions, taking on average more time and tokens. We hypothesize that this increased complexity demands greater model attention and memory, as well as the ability to avoid and correct cumulative errors.

To further explore potential solutions, we tested a subset of 54 non-diagram questions using OpenAI's o1-preview model. This yielded an additional 23 correct answers (42.59\%), demonstrating the accuracy improvement achieved through Chain of Thought (CoT) reasoning embedded in the model. While these results are promising, several limitations hinder broader application. The o1-preview model lacks support for JSON format enforcement, making automatic extraction of answers infeasible. Additionally, it lacks support for encoding images, diagrams, and hyperparameter adjustments, while its higher token cost and processing time presents scalability challenges. Furthermore, as of now, only the preview version of the o1 model has been released by OpenAI, restricting its accessibility for comprehensive testing across the entire dataset.

\begin{remark}\normalfont
    Longer questions present a particular challenge due to higher token requirements and increased complexity. Investigating techniques such as Chain-of-Thought (CoT) prompting, fine-tuning, or length-specific strategies may help models handle these more demanding problems more effectively.
\end{remark}

\subsubsection*{Addressing questions with diagrams}
Diagram-based questions, while showing notable improvements, still have lower accuracy compared to non-diagram questions ($59.57\% (\uparrow n=10)$ vs. $78.66\% (\uparrow n=42)$). The best scenario for diagram questions is to use GPT-4o with translated English questions ($59.57\% (\uparrow n=10)$).
\begin{remark}\normalfont
    Enhancing diagram comprehension in language models is critical. Future work should focus on developing LLM-based diagram comprehension techniques and fine-tuning models using diagram-rich datasets to improve performance in visually intensive question formats.
\end{remark}

\subsection{Future Work}
From our study, we suggest that future works in automated to solve CSAT questions (or broader non-English mathematics tests) should focus on enhancing the comprehension of diagram-based questions in large language models by developing advanced techniques and fine-tuning diagram-rich datasets to address the performance disparity between diagram and non-diagram questions. First, addressing the challenges posed by low-resource languages, such as Korean, will involve fine-tuning models on bilingual datasets and leveraging translation-based methods to improve accuracy in math problem-solving and other tasks. Second, strategies for handling long and complex questions will be explored, including token-efficient techniques like Chain-of-Thought prompting, length-specific processing, and further model fine-tuning. Open-source alternatives, such as Llama-3-70B-Instruct-Gradient, offer flexible and promising avenues for improving model performance and versatility. For closed-source solutions, we recommend the GPT-4o model due to its support for vision, JSON structures, hyperparameter tuning, medium-range pricing, and manageable API limits. Finally, generating high-quality bilingual datasets through high-performing closed-source models like OpenAI’s o1 will serve as a foundation for fine-tuning smaller, open-source models, bridging the gap in performance across diverse scenarios and fostering the development of more robust and inclusive systems.

\section{Conclusion}
To this end, our study highlights both the strengths and limitations of ChatGPT in addressing Korean mathematics problems, offering valuable insights into its potential role in multilingual educational contexts. The model's accuracy in solving these problems (66.72\%) underscores its promising utility in supporting non-English educational settings while simultaneously pointing to significant areas requiring enhancement. These findings suggest that while ChatGPT can handle certain mathematical problems effectively, its performance diminishes for more complex, higher-marked, or visually intensive questions. Notably, ChatGPT’s strong alignment with human annotations in question ratings and categorization reveals its potential for use as an educational tool for assessment design. Its ability to evaluate metrics such as difficulty, cognitive demand, and clarity highlights its applicability to personalized learning systems that cater to diverse student needs. However, its struggles with sequential reasoning and diagram-based questions indicate critical areas for targeted improvements. To fully integrate AI into multilingual and inclusive education, future research should prioritize refining LLMs handling of low-resource languages and complex, multi-step mathematical problems. Enhancing its ability to interpret diagrams and improving accuracy through fine-tuning or adaptive learning approaches will be essential. Furthermore, leveraging advanced models and creating domain-specific datasets can bridge the current performance gaps, setting the stage for more effective and reliable educational AI systems. Ultimately, our study contributes to the growing discourse on the role of AI in global education, advocating for iterative improvements to ensure these technologies meet diverse pedagogical demands. With continued advancements, ChatGPT and similar models hold the potential to transform personalized education, empowering learners worldwide.

\bibliography{references}

\begin{thebibliography}{10}

\bibitem{daher2024use}
W.~Daher and F.~Gierdien.
\newblock Use of language by generative ai tools in mathematical problem solving: The case of chatgpt.
\newblock {\em African Journal of Research in Mathematics, Science and Technology Education}, 28(2):222--235, 2024.

\bibitem{deng2024enhancing}
Y.~Deng, P.~Lu, F.~Yin, Z.~Hu, S.~Shen, J.~Zou, K.-W. Chang, and W.~Wang.
\newblock Enhancing large vision language models with self-training on image comprehension.
\newblock {\em arXiv preprint arXiv:2405.19716}, 2024.

\bibitem{frieder2024mathematical}
S.~Frieder, L.~Pinchetti, R.-R. Griffiths, T.~Salvatori, T.~Lukasiewicz, P.~Petersen, and J.~Berner.
\newblock Mathematical capabilities of chatgpt.
\newblock {\em Advances in neural information processing systems}, 36, 2024.

\bibitem{hsieh2024evaluating}
C.-H. Hsieh, H.-Y. Hsieh, and H.-P. Lin.
\newblock Evaluating the performance of chatgpt-3.5 and chatgpt-4 on the taiwan plastic surgery board examination.
\newblock {\em Heliyon}, 10(14), 2024.

\bibitem{kashefi2023chatgpt}
A.~Kashefi and T.~Mukerji.
\newblock Chatgpt for programming numerical methods.
\newblock {\em Journal of Machine Learning for Modeling and Computing}, 4(2), 2023.

\bibitem{liu2024chatgpt}
J.~Liu, L.~Shen, and G.-W. Wei.
\newblock Chatgpt for computational topology.
\newblock {\em Foundations of data science (Springfield, Mo.)}, 6(2):221, 2024.

\bibitem{mukherjee2008review}
A.~Mukherjee and U.~Garain.
\newblock A review of methods for automatic understanding of natural language mathematical problems.
\newblock {\em Artificial Intelligence Review}, 29:93--122, 2008.

\bibitem{peng2024multimath}
S.~Peng, D.~Fu, L.~Gao, X.~Zhong, H.~Fu, and Z.~Tang.
\newblock Multimath: Bridging visual and mathematical reasoning for large language models.
\newblock {\em arXiv preprint arXiv:2409.00147}, 2024.

\bibitem{siebielec2024assessment}
J.~Siebielec, M.~Ordak, A.~Oskroba, A.~Dworakowska, and M.~Bujalska-Zadrozny.
\newblock Assessment study of chatgpt-3.5’s performance on the final polish medical examination: Accuracy in answering 980 questions.
\newblock In {\em Healthcare}, volume~12, page 1637. MDPI, 2024.

\bibitem{wardat2023chatgpt}
Y.~Wardat, M.~A. Tashtoush, R.~AlAli, and A.~M. Jarrah.
\newblock Chatgpt: A revolutionary tool for teaching and learning mathematics.
\newblock {\em Eurasia Journal of Mathematics, Science and Technology Education}, 19(7):em2286, 2023.

\bibitem{zhang2025mathverse}
R.~Zhang, D.~Jiang, Y.~Zhang, H.~Lin, Z.~Guo, P.~Qiu, A.~Zhou, P.~Lu, K.-W. Chang, Y.~Qiao, et~al.
\newblock Mathverse: Does your multi-modal llm truly see the diagrams in visual math problems?
\newblock In {\em European Conference on Computer Vision}, pages 169--186. Springer, 2025.

\end{thebibliography}
\bibliographystyle{abbrv}

\section*{Appendix}
\subsection*{Statistical test of correlation analysis}
\begin{figure}
    \centering
    \includegraphics[width=\linewidth]{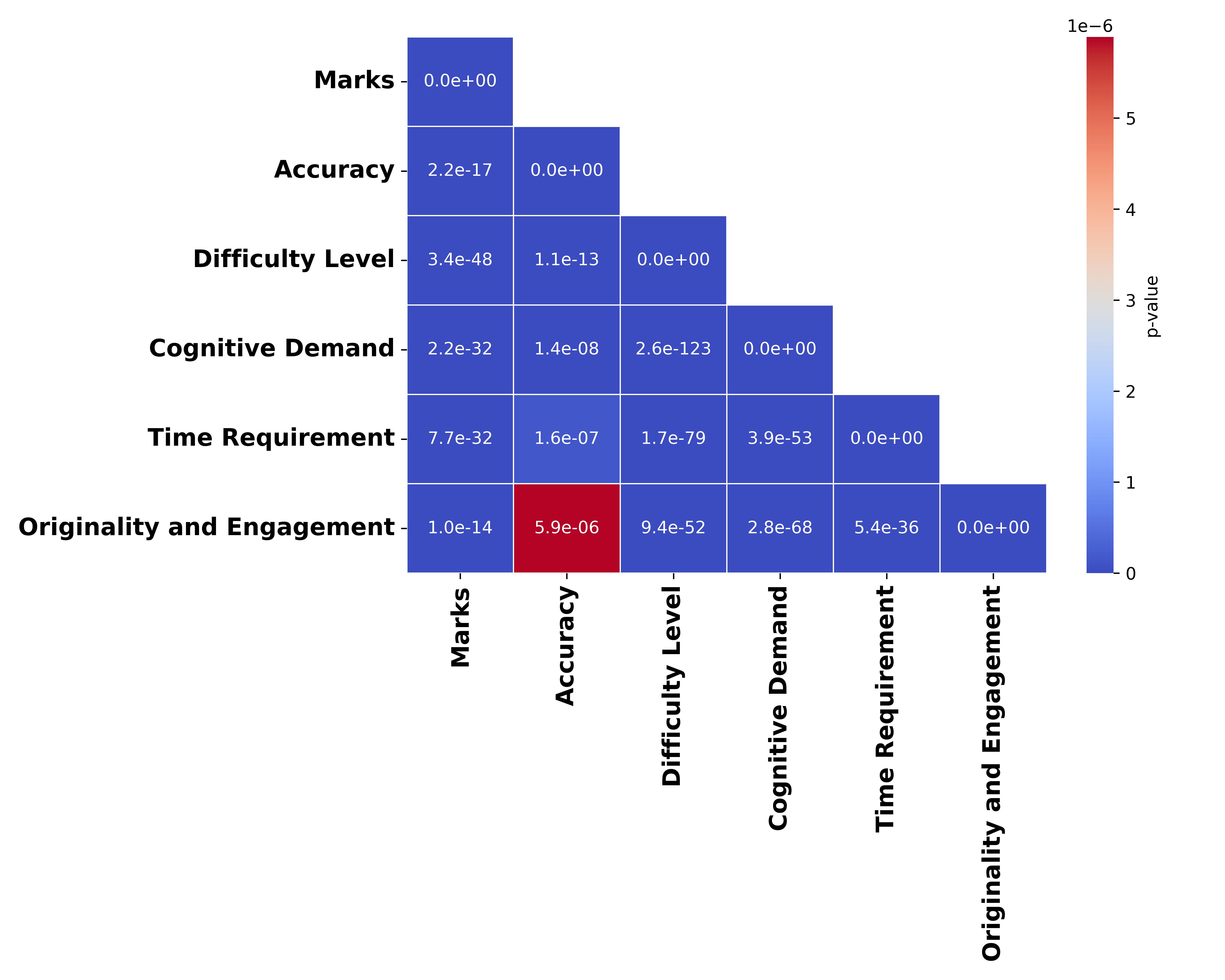}
    \caption{Chi-squared test for the predicted variables in correlation analysis in Figure~\ref{fig:corr_analysis}}
    \label{fig:chisquared_plot}
\end{figure}

\subsection*{Topic distribution by test}

\end{document}